\documentclass[11pt]{article}

\usepackage[preprint]{acl}
\usepackage{times}
\usepackage{latexsym}
\usepackage{multirow}
\usepackage[T1]{fontenc}
\usepackage[table]{xcolor}
\usepackage[utf8]{inputenc}
\usepackage{svg}
\usepackage{microtype}
\usepackage{amssymb}
\usepackage{inconsolata}

\usepackage{graphicx,verbatim}
\usepackage{hyperref}
\usepackage{amsmath}
\usepackage{booktabs}
\usepackage{float}
\usepackage{placeins}
\usepackage{algorithm}
\usepackage{algpseudocode}
\usepackage[most]{tcolorbox}

\title{Latent Ordinal Evidence, Misaligned Outputs: Inference-Time Ordinal Lens Alignment for Multimodal LLMs}

\author{
\textbf{Haiming Li}\textsuperscript{1,2,*}
\quad
\textbf{Yingsheng Liu}\textsuperscript{1,2,*\ensuremath{\dagger}}
\quad
\textbf{Jingmin Zhu}\textsuperscript{1,2,*}
\quad
\textbf{Siyuan Yan}\textsuperscript{1,2}
\\
\textbf{Xieji Li}\textsuperscript{1,2}
\quad
\textbf{Jiajun Sun}\textsuperscript{2}
\quad
\textbf{Zhen Yu}\textsuperscript{1,2,\ensuremath{\ddagger}}
\quad
\textbf{Zongyuan Ge}\textsuperscript{1,2}
\\[4pt]
\textsuperscript{1}Faculty of Information Technology,
Monash University, Melbourne, Australia
\\
\textsuperscript{2}Monash University, Melbourne, Victoria, Australia
}

\begin{document}
\maketitle

\begingroup
\makeatletter
\renewcommand{\thefootnote}{}
\renewcommand{\@makefntext}[1]{\noindent #1}
\footnotetext{%
  \raggedright
  Code: \url{https://anonymous.4open.science/r/OLA-0BEE}\par
  \textsuperscript{*} \textbf{Equal Contribution.}\par
  \textsuperscript{\ensuremath{\dagger}} \textbf{Project Leader.}\par
  \textsuperscript{\ensuremath{\ddagger}} \textbf{Corresponding Author.}
}
\makeatother
\endgroup

\begin{abstract}
Multimodal LLMs apply the language model interface to visual inputs, where ordinal regression tasks such as age estimation, image quality assessment, and disease grading require autoregressive decisions over ordered class labels. We ask whether MLLMs reliably convert internal ordinal evidence into ordered digit-token outputs. Across four ordinal benchmarks and four MLLM backbones, ordinal labels are linearly recoverable from hidden states with Spearman correlation up to 0.938, and a task-designed prompt further sharpens this structure. Yet native digit-token outputs weakly expose it: the unembedding matrix filters the ordinal direction, and the digit-token row space retains below 1.15\% across all 16 model-dataset combinations, with a 16 to 77 absolute-point accuracy gap between linear-probe and native outputs. We introduce Ordinal Lens Alignment (OLA), a frozen-backbone inference-time method that trains lightweight \(W_S\)-anchored lenses on mid-to-deep decoder layers, fuses them into an ordinal distribution, and corrects only digit-token logits at generation. OLA outperforms the SOTA LoRA-tuned OrderChain baseline in most settings while keeping the MLLM frozen, surpasses discriminative ordinal baselines in most cells, and improves over an offline lens in every setting. 
\end{abstract}

\section{Introduction}

\begin{table*}[t]
\centering
\small
\setlength{\tabcolsep}{4pt}
\renewcommand{\arraystretch}{0.8}
\newcolumntype{H}{>{\columncolor{gray!12}}c}  
\begin{tabular}{l cc H | cc H | cc H | cc H}
\toprule
& \multicolumn{3}{c|}{Adience (\(C=8\))} & \multicolumn{3}{c|}{DR (\(C=5\))} & \multicolumn{3}{c|}{HCI (\(C=5\))} & \multicolumn{3}{c}{Aesthetic (\(C=5\))} \\
\cmidrule(lr){2-4} \cmidrule(lr){5-7} \cmidrule(lr){8-10} \cmidrule(lr){11-13}
Backbone & IT/N & LT/N & LT/D & IT/N & LT/N & LT/D & IT/N & LT/N & LT/D & IT/N & LT/N & LT/D \\
\midrule
Qwen3-VL    & 0.916 & 0.872 & \textbf{0.938} & 0.427 & 0.276 & \textbf{0.758} & 0.617 & 0.556 & \textbf{0.843} & 0.859 & 0.864 & \textbf{0.916} \\
Gemma-4     & 0.887 & 0.893 & \textbf{0.929} & 0.362 & 0.255 & \textbf{0.757} & 0.430 & 0.551 & \textbf{0.773} & 0.794 & 0.838 & \textbf{0.919} \\
LLaVA-NeXT  & 0.917 & 0.920 & \textbf{0.924} & 0.421 & 0.365 & \textbf{0.756} & 0.554 & 0.586 & \textbf{0.800} & 0.822 & 0.860 & \textbf{0.906} \\
\bottomrule
\end{tabular}
\caption{SRCC Diagnostics for Probe-Readable Ordinal Evidence}
\label{tab:srcc}
\end{table*}

Ordinal prediction differs from nominal classification because its errors carry distance. Misclassifying age group 3 as group 4 differs qualitatively from misclassifying it as group 7, and similar gradations arise in disease grading, image quality assessment, and aesthetic rating. Discriminative ordinal regression has been addressed through rank-aware decoders \citep{25, 2, 1, 5}, label-smoothing and diversity objectives \citep{8, 7}, domain-specific aesthetic and medical scoring \citep{4, 3}, and vision-language formulations \citep{26, 6, 27}. An alternative route reformulates the task as digit-token generation through prompted MLLMs \citep{28}; this is attractive because it preserves the language-model generation interface and supports rationale-style responses. However, autoregressive token decoding over a discrete vocabulary provides no architectural guarantee that the generated digit tokens respect the ordinal scale of the underlying labels: a language model can emit a valid token without emitting an ordinally appropriate one, a tendency recently audited as central-tendency bias in MLLM clinical ordinal scoring \citep{19}.

\begin{figure}[t]
\centering
\includegraphics[width=0.95\columnwidth]{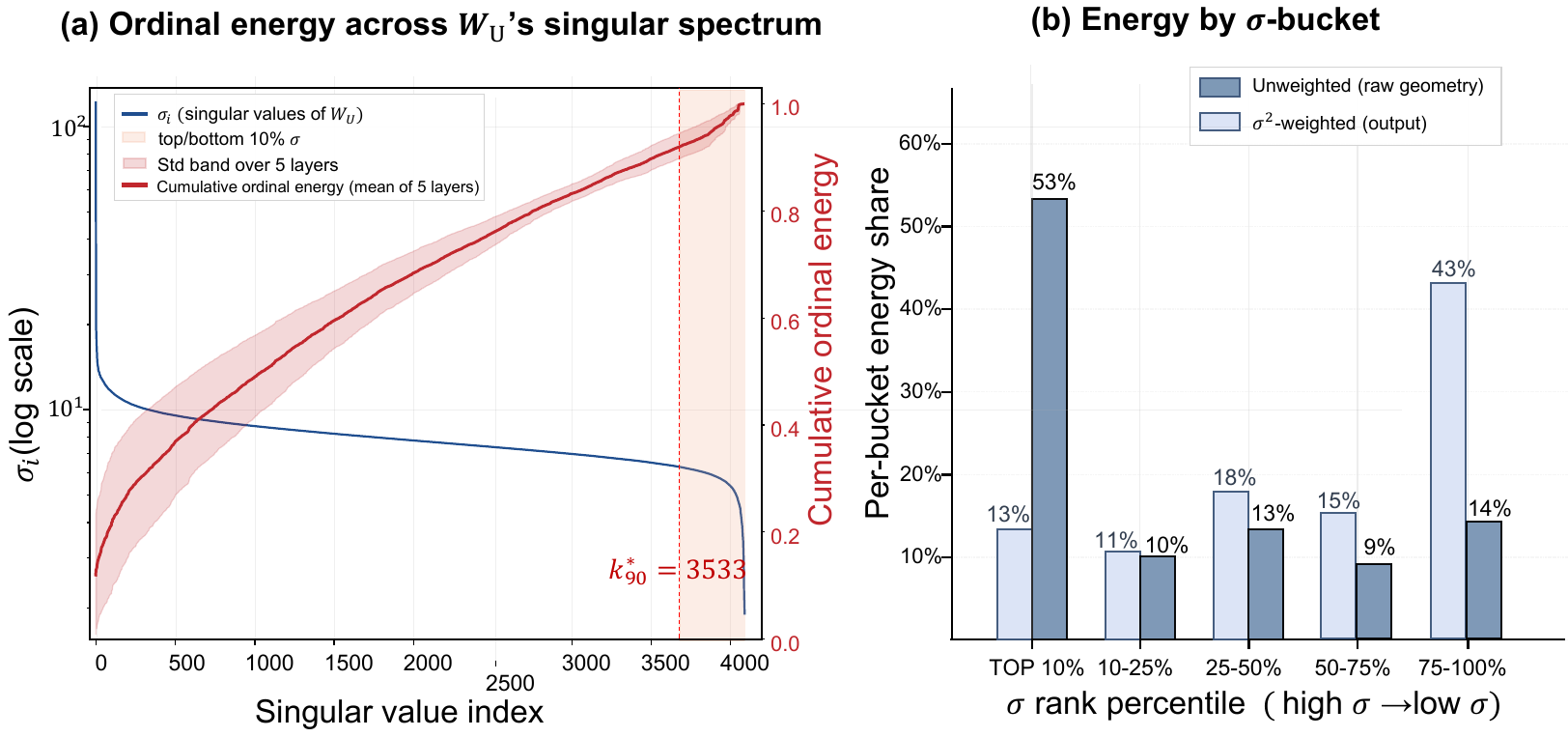}
\caption{Spectral Filtering of the Probed Ordinal Direction by the Unembedding Matrix \(W_U\)}
\label{fig:wu_spectral}
\end{figure}

We examine this issue through a diagnostic study across four ordinal benchmarks (Adience, Diabetic Retinopathy, Historical Color Image, and Aesthetic) and four MLLM backbones (Qwen2.5-VL, Qwen3-VL, Gemma-4, and LLaVA-NeXT). Three observations emerge. First, hidden states contain strong ordinal evidence: linear probes recover ordered labels with peak Spearman correlation up to 0.938 (Table~\ref{tab:srcc}), and a task-designed prompt further sharpens this structure in last-token representations. Second, native digit-token outputs do not reliably expose this evidence: the unembedding matrix \(W_U\) spectrally filters the probed ordinal direction, and the digit-token row space \(W_S\) retains less than 1.15\% of its energy across all 16 model-dataset combinations (Figure~\ref{fig:wu_spectral}, Table~\ref{tab:ws_retention}). Third, this output-side weakness corresponds to a behavioral accuracy gap of 16 to 77 absolute points between linear probes on hidden states and the native digit-token output, indicating that what is linearly decodable internally can differ systematically from what the generation interface produces.

Motivated by this gap, we propose Ordinal Lens Alignment (OLA), a frozen-backbone inference-time method that bridges latent evidence to digit-token logits without modifying the vision encoder, language-model backbone, or unembedding matrix. OLA trains lightweight \(W_S\)-anchored lenses on \(K\) mid-to-deep decoder layers, fuses them via a learned softmax weighting into an instance-specific ordinal distribution \(P^{\text{OLA}}\), and applies a target-restricted, discrepancy-driven correction to digit-token logits at the answer position.

Our contributions are threefold. First, we provide a diagnostic account that separates probe-readable latent ordinal evidence from its weak exposure in native digit-token outputs across four backbones and four benchmarks, with SRCC up to 0.938 yet below 1.15\% \(W_S\) row-space retention in all 16 combinations. Second, we propose Ordinal Lens Alignment (OLA), a frozen-backbone inference-time method that connects multi-layer \(W_S\)-anchored lenses to digit-token logits through a target-restricted, discrepancy-driven correction, without modifying the vision encoder, language-model backbone, or unembedding matrix. Third, OLA improves over prompt-only, hidden-steering, capacity-matched, and offline-lens alternatives, and outperforms the SOTA LoRA-tuned OrderChain baseline in most settings while requiring only lightweight lens training without LoRA fine-tuning.

\section{Task Setup \& Diagnostic Findings}
\label{sec:setup_findings}
\subsection{Task setup}
\label{sec:setup}

We consider visual ordinal prediction with MLLMs. An input \((x, P)\) pairs an image with a prompt specifying the task and answer format; the label \(y\in\{0,1,\ldots,C-1\}\) is ordered, with \(C\) varying across benchmarks. The MLLM predicts via greedy autoregressive decoding; \(\hat y=c\) iff the emitted digit token equals \(s_c\), where \(\mathcal{S}_{\text{digit}}=\{s_0,\ldots,s_{C-1}\}\) is the per-tokenizer digit-token set. Let \(h_\ell(x,P)\in\mathbb{R}^d\) be the last-token hidden state at layer \(\ell\), \(W_U\in\mathbb{R}^{|\mathcal{V}|\times d}\) the unembedding matrix, and \(W_S=W_U[\mathcal{S}_{\text{digit}}]\in\mathbb{R}^{C\times d}\) its digit-token row submatrix. Evaluation uses ACC, MAE, and SRCC: top-1 correctness, ordinal error magnitude, and monotonic agreement of predicted or probed indices with ground-truth labels.

We evaluate on Adience \citep{9} (\(C=8\)), DR \citep{10}, HCI \citep{11}, and Aesthetic \citep{12} (\(C=5\) each) with the open-source MLLM backbones Qwen2.5-VL \citep{13}, Qwen3-VL \citep{14}, Gemma-4 \citep{15}, and LLaVA-NeXT \citep{16}; SRCC diagnostics use the latter three as representatives. Per-layer linear probes on frozen hidden states at three diagnostic sites yield the maximum test SRCC across layers, most recoverable in mid-to-deep layers, suggesting ordinal structure emerges after cross-modal fusion but is weakly exposed at the digit-token interface. Dataset descriptions, class distributions, and probe-site protocols appear in Appendix~\ref{app:datasets}.

\begin{table}[t]
\centering
\small
\setlength{\tabcolsep}{4pt}
\renewcommand{\arraystretch}{0.8}
\begin{tabular}{lcccc}
\toprule
Backbone & Adience & DR & HCI & Aesthetic \\
\midrule
Qwen2.5-VL  & 1.05\% & 0.92\% & 0.99\% & 1.01\% \\
Qwen3-VL    & 1.12\% & 1.06\% & 1.08\% & 1.14\% \\
Gemma-4     & 0.97\% & 0.88\% & 0.93\% & 0.96\% \\
LLaVA-NeXT  & 1.03\% & 0.95\% & 0.97\% & 1.02\% \\
\bottomrule
\end{tabular}
\caption{Digit-Token Row-Space Retention of Probe-Derived Ordinal Directions}
\label{tab:ws_retention}
\end{table}

\subsection{Latent Ordinal Evidence Is Probe-Readable}
\label{sec:finding1}

For each diagnostic site and decoder layer, we train a free linear readout over frozen hidden states, in the spirit of linear probing \citep{17, 18} and lens-based interpretation \citep{24, 21}:
\begin{equation}
\label{eq:probe}
p_{\ell}^{\mathrm{pr}}(c\mid x,P)
=
\mathrm{\sigma}\!\left(B_\ell \tilde h_\ell(x,P)+a_\ell\right)_c
\end{equation}
where $\mathrm{\sigma}(\cdot)$ denotes the softmax and $\tilde h_\ell=(h_\ell-\mu_\ell)/\sigma_\ell$ is standardized. The free probe is unconstrained by native digit-token rows, testing whether ordinal labels are linearly recoverable from frozen representations. Table~\ref{tab:srcc} reports the peak test SRCC across decoder layers under the three protocols of Section~\ref{sec:setup}; per-layer SRCC breakdowns across all three backbones are in Appendix~\ref{app:srcc-per-layer}.

Under the most informative protocol (last-token hidden state with a task-designed prompt), every model-dataset cell reaches SRCC \(\geq 0.756\), peaking at 0.938 on Adience with Qwen3-VL, 0.919 on Aesthetic with Gemma-4, and 0.843 on HCI with Qwen3-VL. Across all 12 backbone-dataset combinations of Table~\ref{tab:srcc}, ordered labels are linearly recoverable from frozen hidden states, so downstream ordinal errors cannot be explained by absent internal evidence.

The task-designed prompt (specifying the ordinal scale and answer format; see Appendix~\ref{app:prompts} for templates and design principles) consistently improves last-token SRCC, with largest gains where the neutral probe is weakest: on DR, SRCC rises from 0.255--0.365 to 0.756--0.758; on HCI, from 0.551--0.586 to 0.773--0.843. Adience and Aesthetic already show strong neutral or visual-token recoverability, so prompt gains are smaller. The prompt thus helps the LM aggregate task-relevant cross-modal evidence at the answer position; whether the native digit-token interface uses it effectively is the focus of the next diagnostic.

\begin{tcolorbox}[colback=blue!4,colframe=blue!50!black,boxsep=2pt,left=4pt,right=4pt,top=2pt,bottom=2pt]
\textbf{Finding 1.} MLLMs contain strong ordinal evidence in hidden states.
\end{tcolorbox}

\subsection{The Output Interface Filters and Weakly Exposes Ordinal Evidence}
\label{sec:finding2}

Finding 1 establishes that ordinal labels are linearly recoverable; we now ask whether the same evidence reaches the digit-token logits. We examine \(W_U\) from two complementary angles, its spectral structure and the digit-token row space \(W_S\), and relate both to the behavioral gap between hidden-state probes and native digit-token outputs.

Let \(W_U = U\Sigma V^\top\) with singular values \(\sigma_1 \geq \cdots \geq \sigma_d\) and right singular vectors \(V_i\). For a probe-derived ordinal direction \(v^{\mathrm{ord}}\), let \(\alpha_i = \langle v^{\mathrm{ord}}, V_i \rangle\). Partitioning singular indices into rank buckets \(\mathcal{B}\) (top 10\%, 10--25\%, 25--50\%, 50--75\%, 75--100\%), we report two normalized energy shares per bucket:
\begin{equation}
\label{eq:bucket_energy}
\begin{aligned}
E^{\mathrm{raw}}(\mathcal{B}) &= \frac{\sum_{i \in \mathcal{B}} \alpha_i^2}{\sum_{j=1}^{d} \alpha_j^2}, \\
E^{\mathrm{out}}(\mathcal{B}) &= \frac{\sum_{i \in \mathcal{B}} \alpha_i^2 \sigma_i^2}{\sum_{j=1}^{d} \alpha_j^2 \sigma_j^2}.
\end{aligned}
\end{equation}

\(E^{\mathrm{raw}}\) measures the raw distribution of the probe direction across the singular spectrum, and \(E^{\mathrm{out}}\) the distribution after the per-direction \(\sigma^2\) weighting that \(v \mapsto W_U v\) imposes. Figure~\ref{fig:wu_spectral} shows the top 10\% bucket carries only 13\% of the raw probe-direction mass yet contributes 53\% of the \(\sigma^2\)-weighted output energy, while the bottom 25\% carries 43\% of the raw mass but only 14\% after \(\sigma^2\) weighting. The output interface therefore reweights the ordinal direction through the singular spectrum of \(W_U\), attenuating a broad low-\(\sigma\) component and amplifying a smaller high-\(\sigma\) one, echoing structured signal patterns previously observed in LLM unembedding spectra \citep{20}.

Since ordinal prediction is evaluated through digit tokens, we measure the retention of \(v^{\mathrm{ord}}\) in the digit-token row space \(W_S\):
\begin{equation}
\label{eq:ws_retention}
R_{W_S}(v^{\mathrm{ord}})=
\frac{\|\Pi_{\mathrm{row}(W_S)}v^{\mathrm{ord}}\|_2^2}{\|v^{\mathrm{ord}}\|_2^2}
\end{equation}
where \(\Pi_{\mathrm{row}(W_S)}\) is the orthogonal projector onto \(\mathrm{row}(W_S)\). Table~\ref{tab:ws_retention} reports \(R_{W_S}\) across four backbones and four benchmarks: cell means span 0.88\%--1.14\%, mean 1.00\%. Only about one percent of the probe-derived ordinal direction is represented in the digit-token row space. This does not imply destruction of the remaining energy or that OLA modifies \(W_S\); the native interface simply exposes this direction weakly. Behaviorally, this filtering aligns with accuracy gaps of 16 to 77 absolute points between hidden-state linear readouts and native digit-token outputs, largest on DR and HCI where native generation is weakest.

\begin{tcolorbox}[colback=blue!4,colframe=blue!50!black,boxsep=2pt,left=4pt,right=4pt,top=2pt,bottom=2pt]
\textbf{Finding 2.} Native digit-token outputs weakly expose the recovered ordinal evidence.
\end{tcolorbox}

Findings 1 and 2 identify an interface-level failure mode: evidence is present in frozen hidden states but only weakly exposed in the native digit-token logits. The intervention should therefore connect probe-readable evidence to digit-token logits at generation, without teaching the backbone or retraining \(W_U\); Section~\ref{sec:method} introduces Ordinal Lens Alignment (OLA) to implement this.

\section{Ordinal Lens Alignment}
\label{sec:method}
Ordinal Lens Alignment (OLA) is a frozen-backbone inference-time procedure that connects probe-readable ordinal evidence in mid-to-deep decoder layers to the digit-token logits at the answer position, without modifying the vision encoder, language-model backbone, or unembedding head. The only learned parameters are lightweight lens and fusion weights, trained once per backbone-dataset pair.

\subsection{Overview}
\label{sec:overview}

\begin{figure*}[t]
\centering
\includegraphics[width=\textwidth]{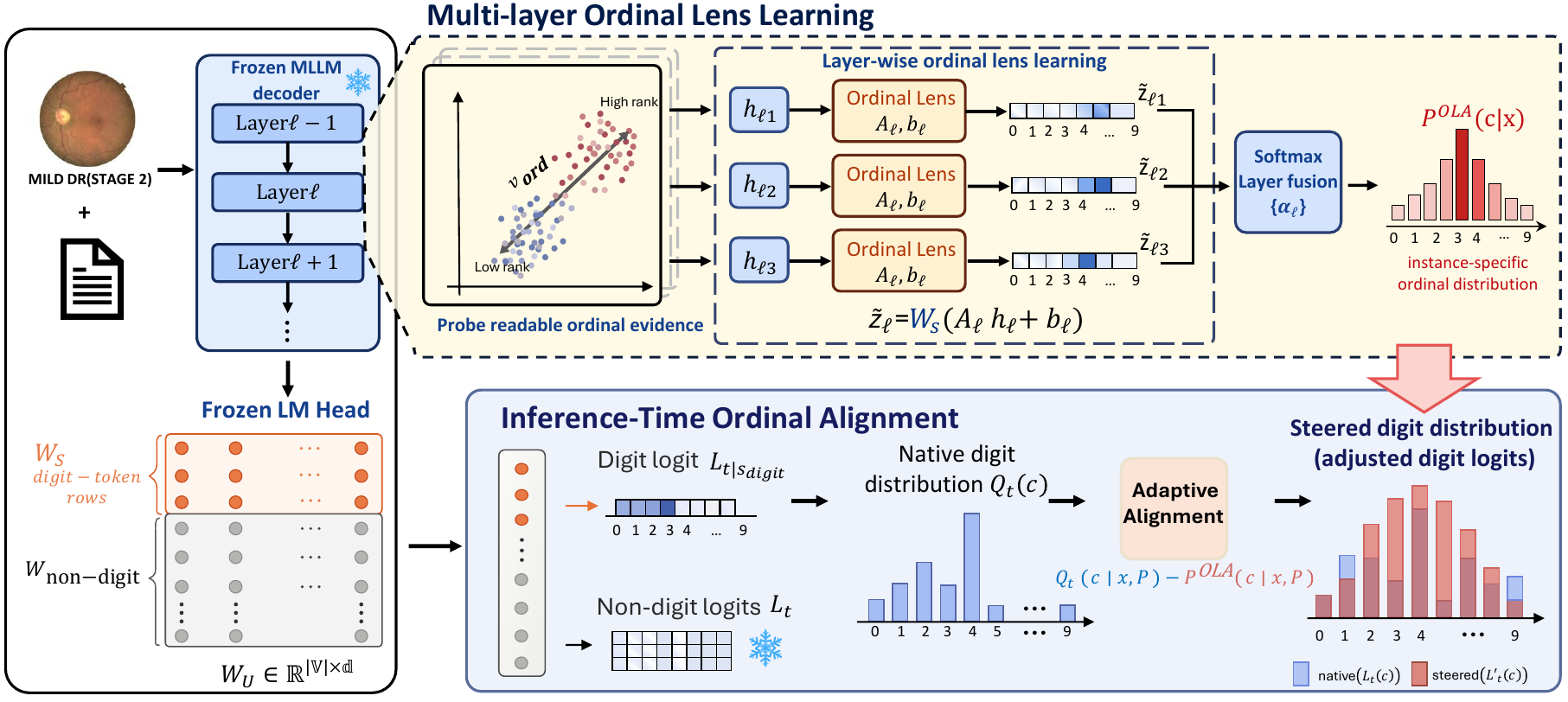}
\caption{Architecture of Ordinal Lens Alignment for Inference-Time}
\label{fig:method}
\end{figure*}

Figure~\ref{fig:method} summarizes OLA as an output-side intervention at the answer-token decoding step. Given \((x,P)\) with task-designed prompt \(P\), the frozen MLLM produces native logits at the answer position. Forward hooks capture last-token hidden states at \(K\) mid-to-deep decoder layers \(\mathcal{L}=\{\ell_1,\ldots,\ell_K\}\); each is passed through a lens mapping it into digit-token scores via the frozen \(W_S\) (Section~\ref{sec:lens}). A learned softmax then fuses per-layer scores into an instance-specific ordinal distribution \(P^{\mathrm{OLA}}(c\mid x,P)\) over \(\mathcal{S}_{\text{digit}}\) (Section~\ref{sec:fusion}). Before greedy decoding selects the answer token, OLA compares \(P^{\mathrm{OLA}}\) with the native digit-token distribution \(Q(c\mid x,P)\) and applies a target-restricted, discrepancy-driven correction only to logits in \(\mathcal{S}_{\text{digit}}\) (Section~\ref{sec:online}); other logits remain unchanged.

This decomposition separates three functions entangled in native MLLM output: latent evidence readout, multi-layer aggregation under a fixed \(K\)-layer window, and inference-time alignment with the digit-token interface. OLA combines logit-lens style readouts \citep{24} with inference-time output-side adjustment, unlike representation steering on hidden states \citep{22} or LoRA-based MLLM adaptation \citep{28}; see Appendix~\ref{app:related_work} for extended discussion.

\subsection{Ordinal Lens}
\label{sec:lens}

The per-layer lens maps a frozen hidden state into a digit-token score vector. Following Findings 1 and 2, it is linear (ordinal labels are linearly recoverable) and anchored on \(W_S\) (native digit-token rows weakly expose the probe direction), expressing probe-readable evidence as digit-token scores rather than repairing \(W_S\) or the LM head.

Let \(h_\ell(x,P)\in\mathbb{R}^{d}\) be the last-token hidden state at layer \(\ell\), captured during the same forward pass as the native logits. For each \(\ell\in\mathcal{L}\), the lens computes
\begin{equation}
\label{eq:lens}
\hat z_\ell(x,P)
=
W_S\bigl(A_\ell \tilde h_\ell(x,P)+b_\ell\bigr)
\end{equation}
with $\tilde h_\ell=(h_\ell-\mu_\ell)/\sigma_\ell$ and $A_\ell=I_d+U_\ell V_\ell^\top$.
Here \(\hat z_\ell\in\mathbb{R}^{C}\) is the digit-token score vector, \(\mu_\ell,\sigma_\ell\in\mathbb{R}^{d}\) are channel-wise statistics on the training split, \(U_\ell,V_\ell\in\mathbb{R}^{d\times r}\) define a low-rank residual transformation (\(r\ll d\)), and \(b_\ell\in\mathbb{R}^{d}\) is a learned offset; trainable parameters are \(\{U_\ell,V_\ell,b_\ell\}\). \(W_S\in\mathbb{R}^{C\times d}\) is frozen, reused from the MLLM unembedding head.

Each layer lens is trained independently with cross-entropy over ordinal labels,
\begin{equation}
\label{eq:lens_loss}
\mathcal{L}_{\mathrm{lens}}^{(\ell)}
=
-\mathbb{E}_{(x,P,y)\sim\mathcal{D}_{\mathrm{train}}}
\log \mathrm{\sigma}\bigl(\hat z_\ell(x,P)\bigr)_y
\end{equation}
on cached hidden states, with no gradient flowing through the MLLM backbone or \(W_S\). Unlike the diagnostic probe with a free matrix \(B_\ell\), the lens factors the readout through \(W_S\), making it an interface module rather than a replacement classifier. Implementation details appear in Appendix~\ref{app:impl}.

\subsection{Aggregation and Online Alignment}
\label{sec:fusion}
\label{sec:online}

The per-layer lens (Section~\ref{sec:lens}) yields \(K\) complementary, \(W_S\)-anchored readouts from different mid-to-deep layers; OLA aggregates them into an ordinal distribution that corrects the native digit-token logits at generation.

OLA fuses the \(K\) per-layer scores via a learned softmax over layers, with \(\alpha\in\mathbb{R}^{K}\) the fusion parameter and \(a=\mathrm{softmax}(\alpha)\in\Delta^{K-1}\) the non-negative weights:
\begin{equation}
\label{eq:fusion}
\begin{aligned}
z^{\mathrm{OLA}}(x,P) &= \sum_{\ell\in\mathcal{L}} a_\ell\,\hat z_\ell(x,P), \\
P^{\mathrm{OLA}}(c\mid x,P) &= \mathrm{\sigma}\bigl(z^{\mathrm{OLA}}(x,P)\bigr)_c.
\end{aligned}
\end{equation}
\(\alpha\) is learned on the training split with lenses frozen via the same cross-entropy objective; the aggregation adds only \(K\) scalars, independent of \(C\) and \(d\).

At inference, the frozen MLLM produces native logit \(L^{\mathrm{na}}(c\mid x,P)\) for digit token \(s_c\), \(c\in\{0,\ldots,C-1\}\); we define the native digit-token distribution \(Q(c\mid x,P)=\mathrm{\sigma}(L^{\mathrm{na}})_c\). Before greedy decoding emits the answer token, OLA applies a discrepancy-driven correction to the digit-token logits:
\begin{equation}
\label{eq:online_correction}
\begin{split}
L'(c\mid x,P)
=&
L^{\mathrm{na}}(c\mid x,P)
\\&-
\lambda^\star\,\omega(P^{\mathrm{OLA}})
\bigl(Q(c\mid x,P)
\\&-P^{\mathrm{OLA}}(c\mid x,P)\bigr)
\end{split}
\end{equation}
for $c\in\{0,\ldots,C-1\}$, where \(\lambda^\star\geq 0\) is selected on the validation split and \(\omega(P^{\mathrm{OLA}})=\max_c P^{\mathrm{OLA}}(c\mid x,P)\) is a confidence gate. The correction is sign-aware: it lowers \(L'(c)\) when \(Q\) assigns more probability to \(c\) than \(P^{\mathrm{OLA}}\) does and raises it otherwise. Logits outside \(\mathcal{S}_{\text{digit}}\) (EOS and non-digit entries) remain unchanged; greedy decoding then emits the ordinal prediction. The complete lens-training and inference-time alignment algorithms appear in Appendix~\ref{app:pseudocode}.

\section{Experiments}
\label{sec:experiments}
This section evaluates OLA along the evidence chain of Section~\ref{sec:setup_findings}. The main benchmark in Section~\ref{sec:main_benchmark} tests whether OLA reduces the behavioral gap between probe-readable hidden-state evidence and native digit-token outputs identified in Section~\ref{sec:finding2}, comparing it with discriminative ordinal references and a LoRA-tuned MLLM baseline. The component ablation in Section~\ref{sec:ablation} isolates which parts of OLA drive the improvement; Section~\ref{sec:analysis} then examines ordinal output structure and hidden-representation geometry. Each result is tied to the diagnostic claim it tests.

\subsection{Main Benchmark}
\label{sec:main_benchmark}

We evaluate four MLLM backbones (Qwen2.5-VL, Qwen3-VL, Gemma-4, LLaVA-NeXT) on the four ordinal benchmarks of Section~\ref{sec:setup}. Adience, DR, and HCI are reported as mean and standard deviation over five folds; Aesthetic uses the balanced test split. For all MLLM-based methods, image inputs, label mappings, digit sets, and evaluation parsers are held consistent within each backbone-dataset setting, while method-specific prompts and training procedures follow the baseline definitions. We compare OLA against three families. The first contains discriminative ordinal references (Ord2Seq, NumCLIP) trained outside the MLLM generation interface. The second contains frozen-backbone MLLM baselines: Naive Prompt for direct digit-token generation under minimal scaffolding, and CAA \citep{22} for validation-selected residual-stream steering with a class-difference direction. The third is OrderChain \citep{28}, the published MLLM ordinal baseline applying LoRA fine-tuning \citep{23} under a task-specific prompt convention. These comparisons span task-specific discriminative prediction, frozen MLLM prompting, hidden-state steering, backbone adaptation, and the output-side alignment OLA introduces.

\begin{table*}[t]
\centering
\small
\setlength{\tabcolsep}{3pt}
\renewcommand{\arraystretch}{0.9}
\begin{tabular}{ll cc cc cc cc c}
\toprule
& & \multicolumn{2}{c}{Adience} & \multicolumn{2}{c}{DR} & \multicolumn{2}{c}{HCI} & \multicolumn{2}{c}{Aesthetic} & \\
\cmidrule(lr){3-4}\cmidrule(lr){5-6}\cmidrule(lr){7-8}\cmidrule(lr){9-10}
Backbone & Method & ACC$\uparrow$ & MAE$\downarrow$ & ACC$\uparrow$ & MAE$\downarrow$ & ACC$\uparrow$ & MAE$\downarrow$ & ACC$\uparrow$ & MAE$\downarrow$ & Mean ACC$\uparrow$ \\
\midrule
\multirow{2}{*}{Discriminative}
 & Ord2Seq      & 0.639 & 0.430 & 0.842 & 0.250 & 0.609 & 0.520 & 0.644 & 0.304 & 0.684 \\
 & NumCLIP      & 0.651 & 0.410 & 0.831 & 0.270 & 0.696 & 0.350 & 0.665 & 0.296 & 0.711 \\
\midrule
\multirow{4}{*}{\shortstack{Qwen2.5-VL\\7B}}
 & Naive Prompt & 0.572 & 0.544 & 0.156 & 1.567 & 0.278 & 0.952 & 0.372 & 0.772 & 0.345 \\
 & CAA          & 0.628 & 0.396 & 0.702 & 0.357 & 0.338 & 0.817 & 0.245 & 0.928 & 0.478 \\
 & OrderChain   & 0.765 & 0.282 & 0.870 & 0.140 & 0.653 & 0.347 & \textbf{0.705} & \textbf{0.295} & 0.748 \\
 & OLA          & \textbf{0.809} & \textbf{0.233} & \textbf{0.920} & \textbf{0.080} & 0.685 & \textbf{0.315} & 0.696 & 0.304 & \textbf{0.778} \\
\midrule
\multirow{4}{*}{\shortstack{Qwen3-VL\\8B}}
 & Naive Prompt & 0.581 & 0.401 & 0.688 & 0.570 & 0.328 & 0.952 & 0.397 & 0.859 & 0.499 \\
 & CAA          & 0.648 & 0.374 & 0.722 & 0.333 & 0.358 & 0.792 & 0.265 & 0.904 & 0.498 \\
 & OrderChain   & 0.790 & 0.210 & 0.895 & 0.110 & 0.655 & 0.530 & \textbf{0.710} & \textbf{0.294} & 0.763 \\
 & OLA          & \textbf{0.806} & \textbf{0.207} & \textbf{0.926} & \textbf{0.078} & \textbf{0.757} & \textbf{0.260} & 0.698 & 0.302 & \textbf{0.797} \\
\midrule
\multirow{4}{*}{\shortstack{Gemma-4\\E4B}}
 & Naive Prompt & 0.571 & 0.489 & 0.137 & 1.429 & 0.245 & 0.996 & 0.214 & 0.963 & 0.292 \\
 & CAA          & 0.574 & 0.437 & 0.744 & 0.307 & 0.319 & 0.896 & 0.341 & 0.816 & 0.495 \\
 & OrderChain   & 0.755 & 0.245 & 0.880 & 0.145 & 0.580 & 0.510 & 0.681 & 0.320 & 0.724 \\
 & OLA          & \textbf{0.760} & \textbf{0.240} & \textbf{0.922} & \textbf{0.078} & \textbf{0.733} & \textbf{0.270} & \textbf{0.701} & \textbf{0.300} & \textbf{0.779} \\
\midrule
\multirow{4}{*}{\shortstack{LLaVA-NeXT\\7B}}
 & Naive Prompt & 0.450 & 0.766 & 0.148 & 1.891 & 0.216 & 1.043 & 0.307 & 0.984 & 0.280 \\
 & CAA          & 0.635 & 0.365 & 0.205 & 1.186 & 0.342 & 0.966 & 0.365 & 0.781 & 0.387 \\
 & OrderChain   & 0.682 & 0.367 & 0.839 & 0.221 & 0.622 & 0.438 & 0.696 & 0.324 & 0.710 \\
 & OLA          & \textbf{0.791} & \textbf{0.229} & \textbf{0.924} & \textbf{0.076} & \textbf{0.723} & \textbf{0.308} & \textbf{0.719} & \textbf{0.281} & \textbf{0.789} \\
\bottomrule
\end{tabular}
\caption{Main benchmark results}
\label{tab:main_results}
\end{table*}

Table~\ref{tab:main_results} reports the main benchmark (full mean$\pm$std over five folds in Appendix~\ref{app:main-results-detailed}). OLA improves over Naive Prompt in every model-dataset setting, with the largest gains on DR and HCI, where the diagnostics in Section~\ref{sec:finding2} show a wide gap between probe-readable hidden-state evidence and weak native digit-token behavior. On DR, OLA raises Qwen2.5-VL from 0.156 to 0.920 ACC, Gemma-4 from 0.137 to 0.922, and LLaVA-NeXT from 0.148 to 0.924; on HCI, OLA raises Qwen3-VL from 0.328 to 0.757 and Gemma-4 from 0.245 to 0.733. These gains do not simply track zero-shot backbone strength; they are strongest where the native output interface underuses recoverable ordinal evidence.

Compared with CAA, OLA is higher in every model-dataset setting. CAA also intervenes on hidden representations but injects a single residual-stream direction and relies on the native LM head to convert that perturbation into a digit-token decision. OLA instead reads ordinal evidence through \(W_S\)-anchored lenses and applies a target-restricted correction directly to digit-token logits. This supports the diagnostic from Section~\ref{sec:finding2}: a strong hidden-state direction alone does not ensure reliable ordinal transfer through the native output interface.

Compared with the SOTA LoRA-tuned OrderChain baseline, OLA achieves higher ACC in 14 of 16 model-dataset settings while keeping the vision encoder, language-model backbone, and unembedding matrix frozen. The two exceptions are Qwen2.5-VL and Qwen3-VL on Aesthetic, where OrderChain is higher by 0.9 and 1.2 ACC points respectively; this boundary case is consistent with the more subjective label structure of aesthetic rating. OLA also exceeds Ord2Seq in all 16 cells and NumCLIP in 15 of 16 cells while preserving the MLLM generation interface. These comparisons indicate that output-side ordinal alignment recovers much of the performance associated with backbone adaptation, without changing frozen MLLM parameters.

Table~\ref{tab:srcc} shows the Designed Prompt sharpens last-token ordinal structure most on DR and HCI. To separate the lens contribution from the prompt contribution, Section~\ref{sec:ablation} holds every trained OLA component fixed while varying only the inference-time prompt.

\subsection{Component Ablation}
\label{sec:ablation}

\begin{table*}[t]
\centering
\footnotesize
\setlength{\tabcolsep}{3pt}
\renewcommand{\arraystretch}{0.9}
\begin{tabular}{ll cc cc cc cc c}
\toprule
& & \multicolumn{2}{c}{Adience} & \multicolumn{2}{c}{DR} & \multicolumn{2}{c}{HCI} & \multicolumn{2}{c}{Aesthetic} & \\
\cmidrule(lr){3-4}\cmidrule(lr){5-6}\cmidrule(lr){7-8}\cmidrule(lr){9-10}
Backbone & Method & ACC$\uparrow$ & MAE$\downarrow$ & ACC$\uparrow$ & MAE$\downarrow$ & ACC$\uparrow$ & MAE$\downarrow$ & ACC$\uparrow$ & MAE$\downarrow$ & Mean ACC$\uparrow$ \\
\midrule
\multirow{8}{*}{\shortstack{Qwen2.5-VL\\7B}}
 & Designed Prompt     & 0.601 & 0.399 & 0.830 & 0.227 & 0.483 & 0.610 & 0.454 & 0.681 & 0.592 \\
 & Probe Steering      & 0.650 & 0.364 & 0.725 & 0.331 & 0.386 & 0.749 & 0.429 & 0.666 & 0.548 \\
 & Probe Only          & 0.780 & 0.267 & 0.910 & 0.095 & 0.623 & 0.390 & 0.670 & 0.350 & 0.746 \\
 & Capacity Probe      & 0.792 & 0.272 & 0.894 & 0.082 & 0.616 & 0.401 & 0.687 & 0.329 & 0.747 \\
 & Single-Layer Logit  & 0.715 & 0.305 & 0.895 & 0.110 & 0.620 & 0.410 & 0.530 & 0.470 & 0.690 \\
 & OL + Naive          & 0.652 & 0.365 & 0.534 & 0.653 & 0.513 & 0.565 & 0.515 & 0.508 & 0.554 \\
 & OLA-Offline         & 0.792 & 0.259 & 0.872 & 0.096 & 0.625 & 0.382 & 0.687 & 0.323 & 0.744 \\
 & OLA-Online          & \textbf{0.809} & \textbf{0.233} & \textbf{0.920} & \textbf{0.080} & \textbf{0.685} & \textbf{0.315} & \textbf{0.696} & \textbf{0.304} & \textbf{0.778} \\
\midrule
\multirow{8}{*}{\shortstack{Qwen3-VL\\8B}}
 & Designed Prompt     & 0.627 & 0.401 & 0.840 & 0.176 & 0.274 & 0.987 & 0.430 & 0.812 & 0.543 \\
 & Probe Steering      & 0.670 & 0.343 & 0.745 & 0.307 & 0.406 & 0.724 & 0.449 & 0.643 & 0.568 \\
 & Probe Only          & 0.785 & 0.230 & 0.911 & 0.089 & 0.690 & 0.356 & 0.620 & 0.440 & 0.752 \\
 & Capacity Probe      & 0.781 & 0.220 & 0.913 & 0.082 & 0.716 & 0.293 & 0.646 & 0.374 & 0.764 \\
 & Single-Layer Logit  & 0.775 & 0.280 & 0.908 & 0.095 & 0.630 & 0.614 & 0.590 & 0.470 & 0.726 \\
 & OL + Naive          & 0.638 & 0.387 & 0.794 & 0.237 & 0.422 & 0.712 & 0.525 & 0.549 & 0.595 \\
 & OLA-Offline         & 0.800 & 0.210 & 0.916 & 0.080 & 0.726 & 0.282 & 0.653 & 0.369 & 0.774 \\
 & OLA-Online          & \textbf{0.806} & \textbf{0.207} & \textbf{0.926} & \textbf{0.078} & \textbf{0.757} & \textbf{0.260} & \textbf{0.698} & \textbf{0.302} & \textbf{0.797} \\
\midrule
\multirow{8}{*}{\shortstack{Gemma-4\\E4B}}
 & Designed Prompt     & 0.692 & 0.309 & 0.828 & 0.179 & 0.292 & 0.902 & 0.348 & 0.855 & 0.540 \\
 & Probe Steering      & 0.623 & 0.387 & 0.785 & 0.258 & 0.322 & 0.880 & 0.333 & 0.826 & 0.516 \\
 & Probe Only          & 0.735 & 0.275 & 0.905 & 0.108 & 0.701 & 0.360 & 0.680 & 0.330 & 0.755 \\
 & Capacity Probe      & 0.732 & 0.261 & 0.913 & 0.082 & 0.729 & 0.315 & 0.682 & 0.316 & 0.764 \\
 & Single-Layer Logit  & 0.725 & 0.290 & 0.895 & 0.120 & 0.650 & 0.410 & 0.620 & 0.430 & 0.723 \\
 & OL + Naive          & 0.684 & 0.320 & 0.493 & 0.705 & 0.392 & 0.772 & 0.563 & 0.512 & 0.533 \\
 & OLA-Offline         & 0.750 & 0.231 & 0.920 & 0.079 & 0.714 & 0.345 & 0.696 & 0.304 & 0.770 \\
 & OLA-Online          & \textbf{0.760} & 0.240 & \textbf{0.922} & \textbf{0.078} & \textbf{0.733} & \textbf{0.270} & \textbf{0.701} & \textbf{0.300} & \textbf{0.779} \\
\midrule
\multirow{8}{*}{\shortstack{LLaVA-NeXT\\7B}}
 & Designed Prompt     & 0.624 & 0.381 & 0.798 & 0.259 & 0.352 & 0.886 & 0.367 & 0.899 & 0.535 \\
 & Probe Steering      & 0.622 & 0.381 & 0.189 & 1.211 & 0.336 & 0.979 & 0.527 & 0.599 & 0.419 \\
 & Probe Only          & 0.755 & 0.302 & 0.840 & 0.175 & 0.702 & 0.350 & 0.705 & 0.301 & 0.751 \\
 & Capacity Probe      & 0.776 & 0.285 & 0.893 & 0.077 & 0.703 & 0.331 & 0.703 & 0.300 & 0.769 \\
 & Single-Layer Logit  & 0.690 & 0.315 & 0.821 & 0.257 & 0.604 & 0.456 & 0.530 & 0.550 & 0.661 \\
 & OL + Naive          & 0.593 & 0.427 & 0.684 & 0.460 & 0.368 & 0.857 & 0.492 & 0.659 & 0.534 \\
 & OLA-Offline         & 0.789 & 0.261 & 0.909 & \textbf{0.071} & 0.718 & 0.329 & 0.715 & \textbf{0.275} & 0.783 \\
 & OLA-Online          & \textbf{0.791} & \textbf{0.229} & \textbf{0.924} & 0.076 & \textbf{0.723} & \textbf{0.308} & \textbf{0.719} & 0.281 & \textbf{0.789} \\
\bottomrule
\end{tabular}
\caption{Component ablation across the design space identified by the diagnostics.}
\label{tab:ablation}
\end{table*}

Table~\ref{tab:ablation} traces OLA across the design space identified by the diagnostics (full mean$\pm$std over five folds in Appendix~\ref{app:ablation-results-detailed}). Free supervised readouts on frozen hidden states (Probe Only, Capacity Probe) already recover much of the gap between Naive Prompt and OLA, confirming Section~\ref{sec:finding1}: the ordinal signal is available in hidden states and does not require a non-linear predictor. Probe Steering, however, stays far below the probe-based classifiers in most settings, even with a supervised probe-derived direction; this supports Section~\ref{sec:finding2}: improving a hidden-state direction alone does not suffice when the intervention still passes through the native output interface.

Comparing Capacity Probe with OLA-Offline tests whether the \(W_S\)-anchored lens and multi-layer aggregation preserve the useful signal while tying the readout to the digit-token interface. OLA-Offline is higher than Capacity Probe in 12 of 16 settings, with an average gain of 0.66 ACC points, useful but not the main source of the final improvement. The most consistent gain comes from reconnecting the fused lens distribution to the native logits: OLA-Online improves over OLA-Offline in all 16 settings (average +1.80 ACC), over Capacity Probe in all 16 (+2.45), and over Probe Only in all 16 (+3.48). These comparisons rule out the view that OLA is only a stronger offline classifier; the decisive step is the target-restricted correction that brings the lens distribution back into the MLLM generation process.

Single-Layer Logit further clarifies the roles of aggregation and online alignment. It outperforms simpler prompt or steering baselines but remains below OLA-Online because it uses only one validation-selected layer and lacks the discrepancy correction of Section~\ref{sec:online}. The full method is thus not just a probe, not just hidden-state steering, and not just single-layer logit substitution; it is an output-side alignment procedure that exploits probe-readable evidence while respecting the digit-token interface identified by the diagnostics.

The OL + Naive row tests prompt dependence directly: the full OLA pipeline trained under the Designed Prompt is applied at inference under the Naive Prompt without re-tuning the lens, fusion, or $\lambda^{\star}$. Across all 16 settings, OL + Naive substantially exceeds the Naive Prompt baseline (e.g., Qwen3-VL on DR: 0.794 vs.\ 0.688; LLaVA-NeXT on DR: 0.684 vs.\ 0.148) while remaining below OLA-Online. The gap to Naive Prompt lower-bounds the share of the OLA gain attributable to prompt-invariant latent ordinal evidence, and the gap to OLA-Online upper-bounds the contribution of the Designed Prompt at inference.

\begin{tcolorbox}[colback=blue!4,colframe=blue!50!black,boxsep=2pt,left=4pt,right=4pt,top=2pt,bottom=2pt]
\textbf{Finding 3.} Target-restricted online alignment reliably converts latent evidence into ordinal decisions.
\end{tcolorbox}

\subsection{Robustness and Qualitative Analyses}
\label{sec:analysis}

Three additional analyses complement the main results: layer-window robustness of the lens, confusion-matrix structure under OLA, and layer-wise PCA of frozen hidden states.

\begin{figure}[t]
\centering
\includegraphics[width=0.9\columnwidth]{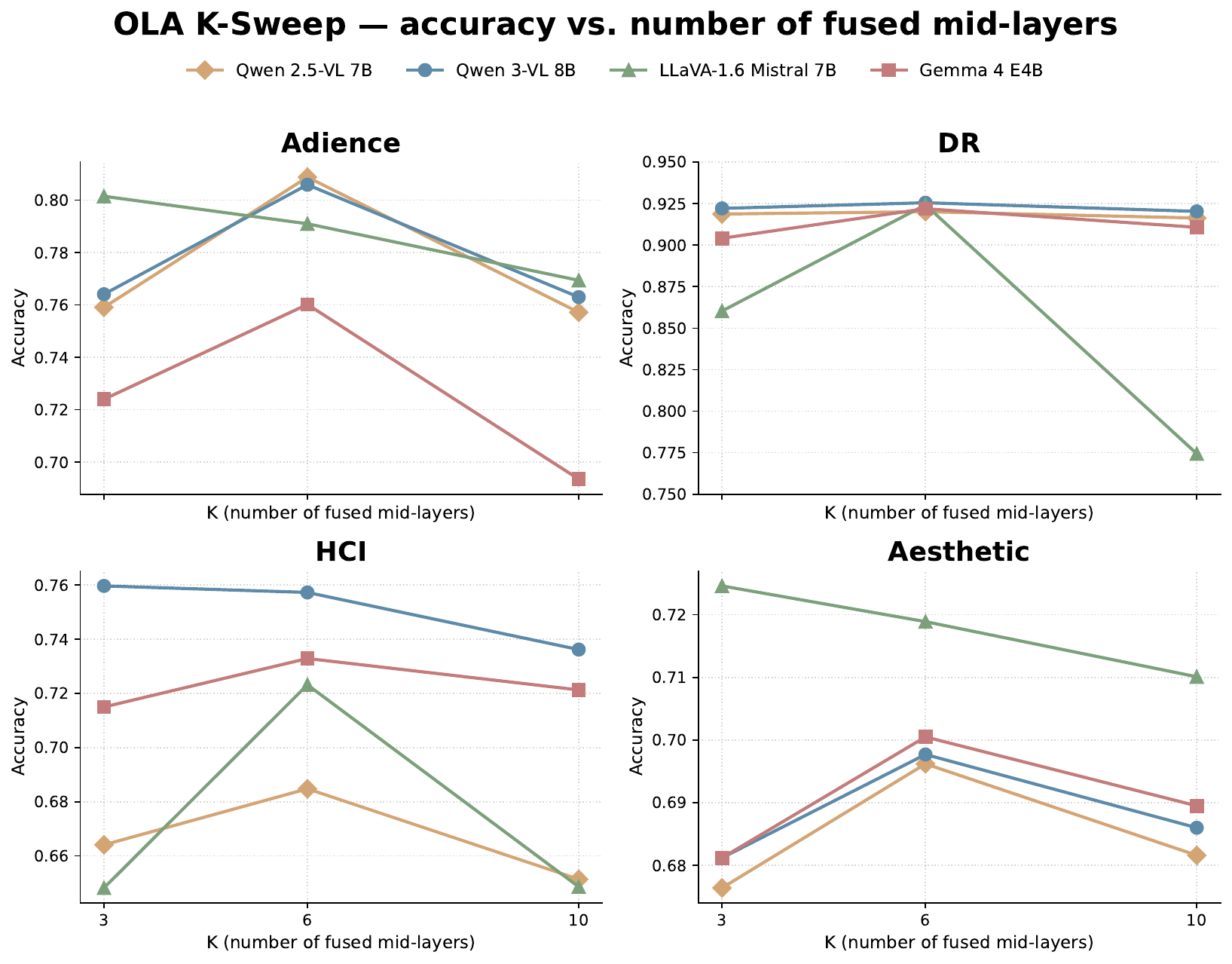}
\caption{OLA accuracy under varying numbers of fused mid-to-deep decoder layers}
\label{fig:k_sweep}
\end{figure}

Figure~\ref{fig:k_sweep} reports OLA accuracy under varying numbers of fused mid-to-deep decoder layers. \(K=6\) is a stable default rather than a universal optimum: in most backbone-dataset curves, \(K=6\) is either best or close to best, while \(K=10\) becomes less stable when the window extends too broadly (most clearly LLaVA-NeXT on DR). Smaller windows are competitive in isolated cases (LLaVA-NeXT on Adience and Aesthetic) but lack the same overall consistency. We fix \(K=6\) throughout the main experiments without selecting \(K\) on the test set, supporting the role of multi-layer aggregation in Section~\ref{sec:fusion}: a stable readout window over the mid-decoder band rather than an oracle tuning mechanism.

\begin{figure}[t]
\centering
\includegraphics[width=0.9\columnwidth]{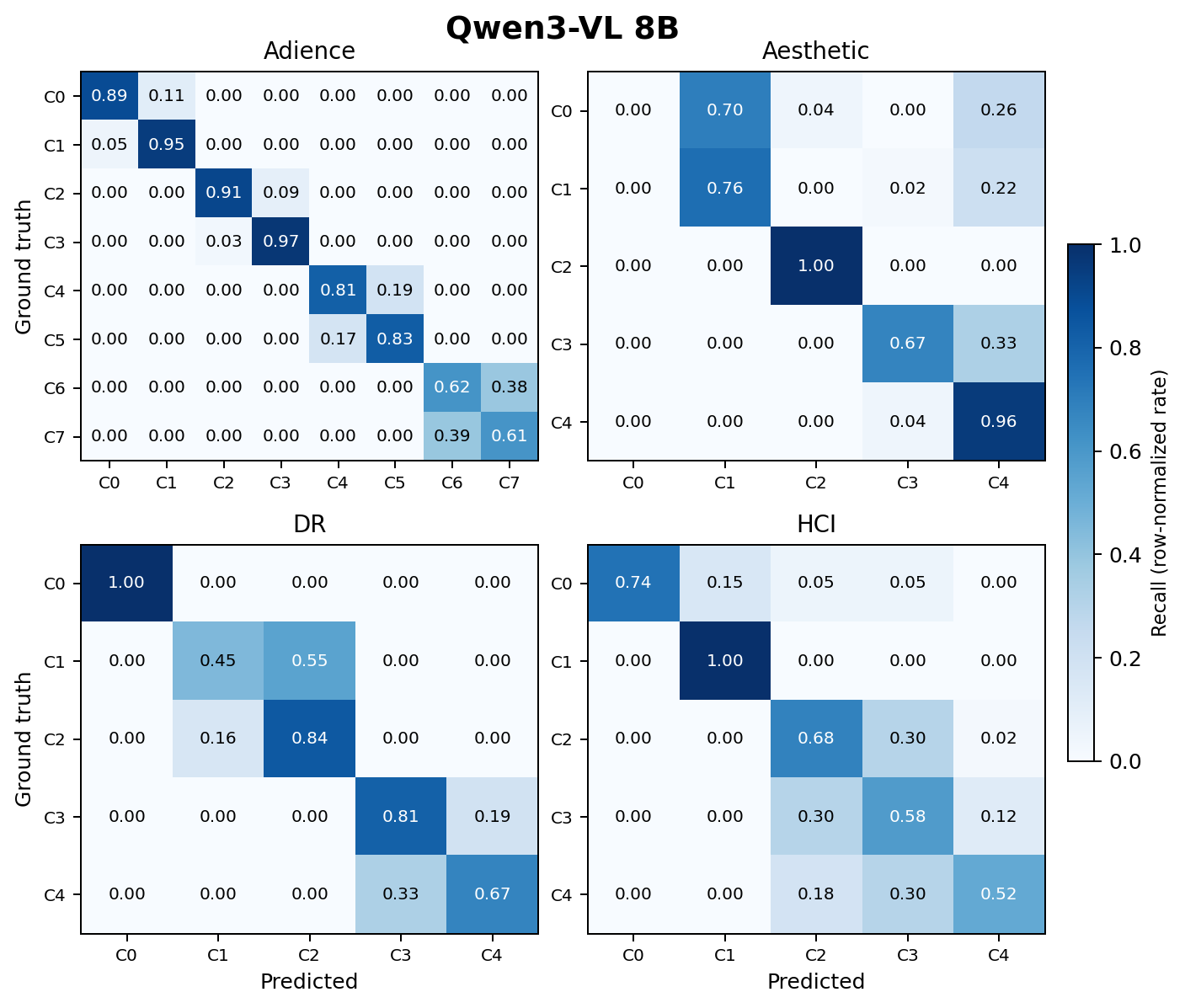}
\caption{Row-Normalized Confusion Matrices}
\label{fig:confusion}
\end{figure}

Figure~\ref{fig:confusion} shows row-normalized confusion matrices of OLA on Qwen3-VL. Adience displays a strong near-diagonal pattern, with most residual errors confined to adjacent age groups. DR and HCI also concentrate errors among nearby ordinal classes, consistent with the MAE reductions in Table~\ref{tab:main_results}. This matters because ordinal prediction is not only a top-1 problem: confusing neighboring classes is less severe than confusing distant ones. Aesthetic remains less diagonal, matching its boundary-case role in the main benchmark and reflecting the noisier structure of subjective ratings. After OLA routes latent ordinal evidence into digit-token logits, residual errors tend to preserve ordinal proximity rather than becoming arbitrary label mistakes.

\begin{figure}[t]
\centering
\includegraphics[width=0.9\columnwidth]{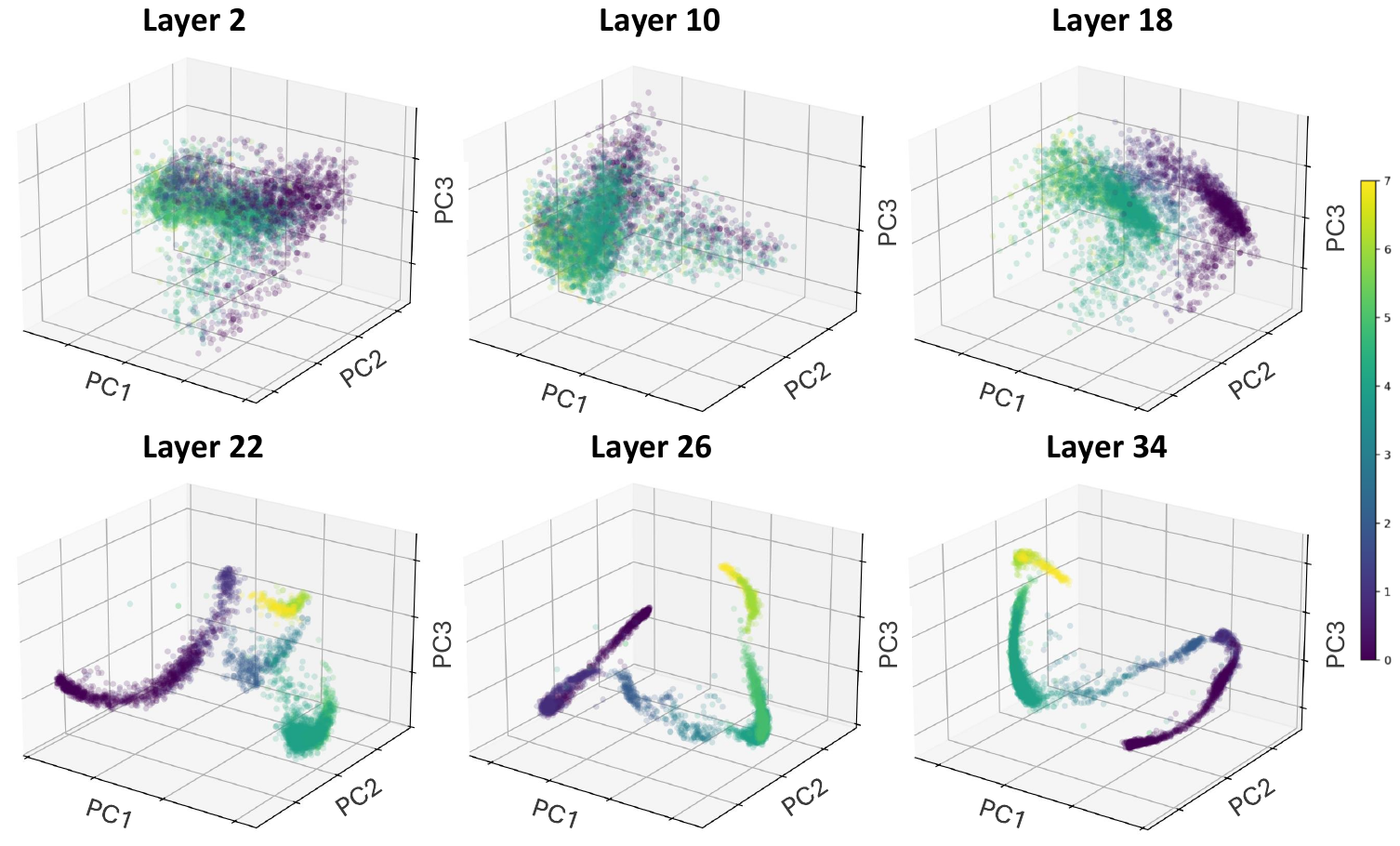}
\caption{Layer-Wise PCA of Last-Token Hidden States}
\label{fig:pca}
\end{figure}

Figure~\ref{fig:pca} visualizes last-token hidden states from Qwen3-VL on Adience across decoder layers. Early layers show substantial class mixing and no clear monotonic color gradient. From the mid-decoder layers onward, the samples organize into a smoother class-ordered structure with a more visible color gradient. This pattern is consistent with the SRCC diagnostic in Table~\ref{tab:srcc}, where ordinal labels become more recoverable from intermediate and deeper hidden states. The visualization should be read qualitatively rather than as causal proof, since PCA only shows a low-dimensional projection; its role is to connect Finding 1 with the layer band used by OLA: the method reads from the part of the frozen MLLM where ordinal information is already organized, and the online correction then exposes it through the digit-token output interface. Cross-backbone PCAs on Adience and cross-dataset PCAs on Qwen3-VL appear in Appendix~\ref{app:pca-per-layer}.

\section{Conclusion}
\label{sec:conclusion}
Across four ordinal benchmarks and four MLLM backbones, latent ordinal evidence is linearly recoverable from frozen hidden states, yet the digit-token row space exposes less than 1.15\% of this energy. OLA bridges the gap with \(W_S\)-anchored lenses and a target-restricted online correction, matching or exceeding LoRA-tuned baselines without modifying the frozen MLLM.

\section*{Limitations}

We highlight three scope conditions of the present study.

First, our experiments cover four open-source MLLM backbones (Qwen2.5-VL, Qwen3-VL, Gemma-4, LLaVA-NeXT) and four ordinal benchmarks. Closed-source MLLMs (e.g., GPT-4V, Claude, Gemini) are not evaluated, since OLA requires access to mid-to-deep hidden states and the digit-token row submatrix \(W_S\). The method applies directly to open-source models with standard transformer interfaces; extending to API-based models requires comparable logit access.

Second, all four benchmarks use single-digit ordinal label tokens (\(C\leq 8\)). Tasks with multi-digit ordinal outputs (e.g., direct year prediction, multi-digit severity scales) would require either label-tokenizer redesign or extending the discrepancy correction to token sequences. We view this as an orthogonal extension rather than a fundamental obstacle.

Third, OLA trains lightweight lens and fusion parameters per backbone-dataset pair (lens: \(\mathcal{O}(Kdr)\) parameters; fusion: \(K\) scalars). Training operates on cached hidden states without backbone gradients, so the cost is much smaller than LoRA fine-tuning, but the method is not fully training-free. Lens transfer across datasets within the same backbone is left to future work.


\bibliography{custom}

\clearpage
\appendix

\section{Datasets}
\label{app:datasets}

This section documents the four ordinal benchmarks used in the main paper. For each dataset we give a brief overview that lists the source publication, the number of ordinal classes, and the per-class sample count, followed by the specific experimental settings.

\subsection{Adience}
Adience \citep{9} is a benchmark for unfiltered face-image age estimation. The ordinal label space partitions human age into \(C=8\) ordered groups (0--2, 4--6, 8--13, 15--20, 25--32, 38--43, 48--53, \(60+\) years); boundaries are non-uniform in calendar years, which makes Adience a canonical benchmark for ordinal classification rather than ordinary multi-class age prediction. We follow the official five-fold protocol released with Adience. Per-class counts are reported in Table~\ref{tab:adience-distribution}.

\begin{table}[!h]
\centering
\resizebox{\columnwidth}{!}{%
\begin{tabular}{@{}lrrrrrrrr@{}}
\toprule
\(y\)         & 0     & 1     & 2     & 3     & 4     & 5     & 6   & 7   \\
Range (yrs)   & 0--2  & 4--6  & 8--13 & 15--20 & 25--32 & 38--43 & 48--53 & 60+ \\
\midrule
Count         & 2{,}488 & 2{,}140 & 2{,}124 & 1{,}642 & 5{,}081 & 2{,}339 & 830 & 872 \\
\bottomrule
\end{tabular}}
\caption{Adience class distribution.}
\label{tab:adience-distribution}
\end{table}

\subsection{Diabetic Retinopathy (DR)}
The Diabetic Retinopathy Detection corpus \citep{10} contains 35{,}126 retinal-fundus photographs graded on the five-level International Clinical Diabetic Retinopathy (ICDR) scale: \(0\)~=~no DR, \(1\)~=~mild non-proliferative DR (NPDR), \(2\)~=~moderate NPDR, \(3\)~=~severe NPDR, \(4\)~=~proliferative DR. We adopt a random five-fold split (75\%/5\%/20\% for training, validation, and test), giving 28{,}101 training-plus-validation images and 7{,}025 test images per fold.

\begin{table}[!h]
\centering
\resizebox{\columnwidth}{!}{%
\begin{tabular}{@{}lrrrrr@{}}
\toprule
\(y\)         & 0     & 1     & 2     & 3      & 4   \\
Grade         & No DR & Mild  & Mod.  & Severe & Prolif. \\
\midrule
Count         & 25{,}810 & 2{,}443 & 5{,}292 & 873 & 708 \\
\bottomrule
\end{tabular}}
\caption{DR class distribution.}
\label{tab:dr-distribution}
\end{table}

\subsection{Historical Color Images (HCI)}
The Historical Color Image dataset \citep{11} is a 1{,}325-image benchmark for historical-photograph dating. Each photograph is annotated with the decade in which it was taken, giving \(C=5\) ordered classes (1930s, 1940s, 1950s, 1960s, 1970s). The corpus is exactly class-balanced (265 images per decade). We adopt a random five-fold split (75\%/5\%/20\% for training, validation, and test).

\subsection{Aesthetic}
The Aesthetic dataset \citep{12} is a 12{,}984-image collection covering four themes (Nature, Animal, Urban, People), annotated on a five-level scale that maps to the ordinal labels \(0\)~=~unacceptable, \(1\)~=~flawed, \(2\)~=~ordinary, \(3\)~=~professional, \(4\)~=~exceptional. Because the corpus is sharply imbalanced across classes, we adopt a class-balanced test partition.

\begin{table}[!h]
\centering
\resizebox{\columnwidth}{!}{%
\begin{tabular}{@{}lrrrrr@{}}
\toprule
\(y\)         & 0     & 1     & 2     & 3     & 4 \\
Quality       & Unaccept. & Flawed & Ordinary & Profess. & Except. \\
\midrule
Count         & 23 & 552 & 7{,}603 & 4{,}574 & 232 \\
\bottomrule
\end{tabular}}
\caption{Aesthetic class distribution.}
\label{tab:aesthetic-distribution}
\end{table}

\subsection{Cross-dataset summary}
Table~\ref{tab:dataset-summary} consolidates the four datasets' ordinal class count, corpus size, per-fold test size, fold count, and split-protocol family.

\begin{table}[H]
\centering
\resizebox{\columnwidth}{!}{%
\begin{tabular}{@{}llcccc@{}}
\toprule
Dataset & Attribute & \(C\) & Train+val & Test/fold & Folds \\
\midrule
Adience    & age              & 8 & \(\sim 14{,}900\) & \(\sim 3{,}700\) & 5 (official) \\
DR         & disease grade    & 5 & 28{,}101       & 7{,}025       & 5 (random) \\
HCI        & decade           & 5 & 1{,}060        & 265           & 5 (random) \\
Aesthetic  & aesthetic rating & 5 & 12{,}777       & 207           & Balanced test \\
\bottomrule
\end{tabular}}
\caption{Per-dataset split summary.}
\label{tab:dataset-summary}
\end{table}

\subsection{Probe-Site Protocol}
To probe latent ordinal evidence, we extract hidden states at three diagnostic sites: (i) the mean of visual placeholder tokens with a neutral caption prompt, (ii) the assistant-start position under the same neutral prompt, and (iii) the assistant-start position under a task-designed prompt that specifies the ordinal scale and answer format. For each backbone, dataset, and diagnostic site, we fit per-layer linear probes on the training split and report the maximum test SRCC across layers as the diagnostic summary.

\FloatBarrier

\section{Prompt Templates}
\label{app:prompts}

This section presents the design principles of the Designed Prompts used to elicit ordinal answers from the four MLLM backbones. The four dataset-specific templates differ only in their domain content; they share the same four-component skeleton.

\subsection{Design principles}
Every Designed Prompt is composed of four components delivered in the following order: an AI role, a domain-knowledge background, an explicit task statement, and a coarse-to-fine reasoning step (CRD). The role primes the backbone with a domain-relevant persona; the background grounds the answer in the precise ordinal label scale; the task statement fixes the answer format; and the CRD step is the conditional chain-of-thought that the lens later reads from the hidden states.

Component (i), the AI role, is a one-sentence persona that primes the backbone with a domain-relevant identity. Component (ii), domain-knowledge background, is a short factual statement that enumerates the ordinal label scale of the dataset, including the semantic anchor of each class. Component (iii), the task statement, is a direct instruction that tells the backbone what to predict and in what format (a single digit, wrapped in a dataset-specific answer delimiter such as \texttt{\$N\$} or \texttt{\&N\&}). Component (iv), coarse-to-fine reasoning (CRD), is a two-step conditional chain that asks the backbone to first commit to a coarse super-class (a contiguous group of adjacent ordinal classes), and then commit to a fine class within that super-class. The coarse partition is a fixed, dataset-specific hierarchy that we define once and reuse across all four backbones; it is not derived from any test-time information, so the Designed Prompt contains no label leakage.

The four components together form a prompt that aligns the backbone's reasoning trace with the ordinal geometry that the lens reads at inference time. Empirically, replacing this prompt with the Naive Prompt (a bare class-mapping question without role, background, or coarse-to-fine reasoning) reduces last-token ordinal SRCC substantially in the deep layers, which is the diagnostic we report in the main paper.

\subsection{Example template: Diabetic Retinopathy}
The following Designed Prompt for the Diabetic Retinopathy (DR) dataset illustrates the four-component skeleton. The remaining three templates (Adience, HCI, Aesthetic) follow the same structure with domain-appropriate substitutions.

\textbf{Role.} You are an ophthalmologist trained to grade diabetic retinopathy from colour fundus photographs on the ICDR scale.
\textbf{Background.} The ICDR scale defines five severity grades: \(0\) denotes no diabetic lesions; \(1\) mild NPDR with microaneurysms only; \(2\) moderate NPDR; \(3\) severe NPDR; \(4\) proliferative DR with neovascularisation.
\textbf{Task.} Grade the DR severity in the fundus photograph and output a single digit from \(0\) to \(4\) wrapped as \texttt{\&N\&}.
\textbf{Coarse-to-fine reasoning.} First commit to one of three coarse categories: Healthy for grade~\(0\); Non-proliferative DR for grades \(1\) to \(3\); Proliferative DR for grade~\(4\). State the visible lesion types that justify the coarse choice. If the coarse category is Non-proliferative DR, refine to mild~\(1\), moderate~\(2\), or severe~\(3\) based on lesion count and distribution. Otherwise commit directly to \(0\) or \(4\). Output the final answer as \texttt{\&N\&}.

\section{Experimental Setup}
\label{app:impl}

This section consolidates the experimental settings for the two trained components that the appendix relies on: the per-layer SRCC probe used in the diagnostic of Appendix~\ref{app:extended-results}, and the multi-layer ordinal lens used by OLA at inference.

\subsection{SRCC probe}
\label{app:srcc-setup}

We extract the last-token hidden state from the frozen base backbone under the Designed Prompt with image input, $L_2$-normalise it, and fit a linear probe $\text{Linear}(d \to 1)$ per layer with mean-squared-error loss against the ordinal label. The optimiser is Adam with learning rate $10^{-4}$, with cosine annealing over $100$ epochs, batch size $128$, and a fixed seed of $42$. The reported metric is the Spearman rank-correlation between the probe's scalar prediction and the ordinal label. The probe head is selected by best test SRCC under a val-as-test protocol on Adience, DR, and Aesthetic; HCI uses its native validation split. Three backbones are probed: Qwen3-VL, Gemma-4, and LLaVA-NeXT.

\subsection{Ordinal lens training}
\label{app:lens-setup}

Training is in two stages, both with the MLLM backbone, the unembedding $W_U$, and its digit-token submatrix $W_S$ strictly frozen; $W_S$ integrity is verified by a SHA-256 hash before and after each stage. Inputs are the last-token hidden states cached from a single forward pass of the frozen MLLM at the $K{=}6$ mid-decoder layers $\mathcal{L}$. Per-layer channel-wise statistics $(\mu_\ell, \sigma_\ell)$ are computed once on the training split with $\sigma_\ell$ clamped to a floor of $10^{-3}$, and the hidden state is standardised as $\tilde h_\ell = (h_\ell - \mu_\ell)/\sigma_\ell$ throughout.

\paragraph{Stage A: per-layer ordinal lens.}
For each layer $\ell \in \mathcal{L}$ we independently train the lens parameters $\{U_\ell, V_\ell, b_\ell\}$ that define $A_\ell = I + U_\ell V_\ell^{\top}$ with rank $r{=}64$. $U_\ell$ is initialised from $\mathcal{N}(0, 0.02^{2})$ and $V_\ell$ is initialised to zero, so the initial transformation collapses to the identity. The objective is cross-entropy of $\mathrm{\sigma}(W_S(A_\ell \tilde h_\ell + b_\ell))$ against the ordinal label. The optimiser is AdamW with learning rate $10^{-4}$, batch size $128$, gradient-clip norm $1.0$, and a per-layer seed of $42 + \mathrm{layer\_id}$. Each layer trains for at most $100$ epochs and the best-epoch parameters are kept under validation accuracy.

\paragraph{Stage B: softmax fusion.}
With all $(A_\ell, b_\ell)$ frozen, we train the $K$ fusion logits $\alpha \in \mathbb{R}^K$ such that $a_\ell = \mathrm{\sigma}(\alpha)_\ell$. The objective is cross-entropy of $\mathrm{\sigma}(\sum_\ell a_\ell\, W_S(A_\ell \tilde h_\ell + b_\ell))$ against the ordinal label. The optimiser is AdamW with learning rate $10^{-2}$, no weight decay, batch size $128$, and seed $42$. Stage B trains for at most $30$ epochs and the best-epoch $\alpha$ is kept under validation accuracy. After Stage B we cache $P^{\mathrm{OLA}}(\cdot \mid x)$ on the train, validation, and test splits for downstream phases.

\onecolumn

\section{Extended Experimental Results}
\label{app:extended-results}

This section reports a per-layer SRCC diagnostic across three MLLM backbones and the full main-benchmark and ablation tables with five-fold standard deviations.

\subsection{Per-layer SRCC across backbones and datasets}
\label{app:srcc-per-layer}

We probe each layer of three MLLM backbones on all four datasets following the protocol in Appendix~\ref{app:srcc-setup}, with Qwen3-VL having $36$ decoder layers and hidden $4096$, Gemma-4 having $42$ layers and hidden $2560$, and LLaVA-NeXT having $32$ layers and hidden $4096$.

\begin{table}[H]
\centering\small
\setlength{\tabcolsep}{8pt}
\begin{tabular}{@{}ccccc@{}}
\toprule
Layer & Adience & DR & HCI & Aesthetic \\
\midrule
 2 & 0.6845 & 0.7522 & 0.5771 & 0.7405 \\
 4 & 0.7283 & 0.7433 & 0.3927 & 0.7175 \\
 6 & 0.7154 & 0.7503 & 0.4278 & 0.7301 \\
 8 & 0.7423 & 0.7530 & 0.4769 & 0.7607 \\
10 & 0.7259 & 0.7539 & 0.4980 & 0.7773 \\
12 & 0.7632 & 0.7542 & 0.5234 & 0.8155 \\
14 & 0.7966 & 0.7557 & 0.5467 & 0.8214 \\
16 & 0.8589 & 0.7564 & 0.6977 & 0.8568 \\
18 & 0.8989 & 0.7577 & 0.7627 & 0.9106 \\
20 & \textbf{0.9376} & \textbf{0.7578} & 0.8148 & 0.9140 \\
22 & 0.9353 & 0.7578 & 0.8377 & 0.9144 \\
24 & 0.9344 & 0.7578 & 0.8235 & \textbf{0.9163} \\
26 & 0.9336 & 0.7578 & 0.8342 & 0.9145 \\
28 & 0.9337 & 0.7578 & \textbf{0.8432} & 0.9127 \\
30 & 0.9342 & 0.7578 & 0.8271 & 0.9114 \\
34 & 0.9347 & 0.7564 & 0.7983 & 0.9051 \\
\bottomrule
\end{tabular}
\caption{Per-layer SRCC on Qwen3-VL.}
\label{tab:srcc-qwen3vl}
\end{table}

SRCC peaks in the upper-middle decoder (L20--L28) and the peaks on Adience and Aesthetic fall inside the layer window \(\mathcal{L} = \{16, 18, 20, 22, 24, 26\}\) used by OLA.

\begin{table}[H]
\centering\small
\setlength{\tabcolsep}{8pt}
\begin{tabular}{@{}ccccc@{}}
\toprule
Layer & Adience & DR & HCI & Aesthetic \\
\midrule
 2 & 0.7987 & 0.7485 & 0.3744 & 0.4907 \\
 9 & 0.8023 & 0.7492 & 0.2497 & 0.5670 \\
12 & 0.8095 & 0.7541 & 0.2413 & 0.7309 \\
15 & 0.8481 & 0.7554 & 0.5146 & 0.7909 \\
18 & 0.8866 & 0.7564 & 0.7188 & 0.7934 \\
20 & 0.8885 & 0.7563 & 0.7251 & 0.8610 \\
21 & 0.8927 & \textbf{0.7567} & 0.7140 & 0.8662 \\
22 & 0.9058 & 0.7565 & 0.7458 & 0.8874 \\
24 & 0.9252 & 0.7560 & 0.7512 & 0.8960 \\
26 & 0.9245 & 0.7562 & \textbf{0.7728} & 0.9083 \\
28 & 0.9166 & 0.7561 & 0.7224 & 0.8748 \\
33 & 0.9210 & 0.7559 & 0.6395 & 0.8864 \\
36 & 0.9262 & 0.7561 & 0.6477 & 0.8976 \\
40 & \textbf{0.9293} & 0.7563 & 0.6962 & \textbf{0.9192} \\
\bottomrule
\end{tabular}
\caption{Per-layer SRCC on Gemma-4.}
\label{tab:srcc-gemma4}
\end{table}

The Gemma-4 decoder peaks later than Qwen3-VL (L40 on Adience and Aesthetic, L26 on HCI), with DR plateauing at \(\rho \approx 0.756\).

\begin{table}[H]
\centering\small
\setlength{\tabcolsep}{8pt}
\begin{tabular}{@{}ccccc@{}}
\toprule
Layer & Adience & DR & HCI & Aesthetic \\
\midrule
 2 & 0.7751 & 0.7495 & 0.3611 & 0.6679 \\
 4 & 0.8164 & 0.7487 & 0.3857 & 0.7171 \\
 6 & 0.8304 & 0.7504 & 0.3736 & 0.6762 \\
 8 & 0.8504 & 0.7504 & 0.3742 & 0.6867 \\
10 & 0.8608 & 0.7525 & 0.3568 & 0.6788 \\
12 & 0.8666 & 0.7557 & 0.3250 & 0.6950 \\
14 & 0.8950 & \textbf{0.7559} & 0.7214 & 0.8930 \\
16 & 0.9164 & 0.7554 & 0.7960 & \textbf{0.9056} \\
18 & 0.9149 & 0.7548 & 0.7921 & 0.8983 \\
20 & 0.9136 & 0.7544 & 0.7784 & 0.8935 \\
22 & 0.9152 & 0.7549 & 0.7891 & 0.9034 \\
24 & 0.9163 & 0.7553 & 0.7974 & 0.9021 \\
26 & \textbf{0.9236} & 0.7557 & 0.7960 & 0.8978 \\
28 & 0.9216 & 0.7556 & 0.7961 & 0.8947 \\
30 & 0.9192 & 0.7556 & \textbf{0.8003} & 0.9056 \\
\bottomrule
\end{tabular}
\caption{Per-layer SRCC on LLaVA-NeXT.}
\label{tab:srcc-llava-next}
\end{table}

LLaVA-NeXT carries a non-trivial ordinal signal already at L2 (\(\rho \ge 0.66\) on three of four datasets) and peaks around L16--L30 across datasets. Table~\ref{tab:srcc-best} summarises the best layer per backbone-dataset cell: the best-layer SRCC falls in \([0.756, 0.938]\) across all 12 cells and the best layer is consistently in the upper-middle of the decoder, never the final layer.

\begin{table}[H]
\centering\small
\setlength{\tabcolsep}{10pt}
\begin{tabular}{@{}lcccc@{}}
\toprule
Backbone $\backslash$ Dataset & Adience & DR & HCI & Aesthetic \\
\midrule
Qwen3-VL    & L20 / \textbf{0.9376} & L20 / 0.7578 & L28 / 0.8432 & L24 / 0.9163 \\
Gemma-4     & L40 / 0.9293          & L21 / 0.7567 & L26 / 0.7728 & L40 / 0.9192 \\
LLaVA-NeXT  & L26 / 0.9236          & L14 / 0.7559 & L30 / 0.8003 & L16 / 0.9056 \\
\bottomrule
\end{tabular}
\caption{Best layer and best SRCC per backbone--dataset cell.}
\label{tab:srcc-best}
\end{table}

\FloatBarrier

\subsection{Layer-wise 3D PCA visualisations}
\label{app:pca-per-layer}

\begin{figure}[!htbp]
\centering
\includegraphics[width=0.85\linewidth]{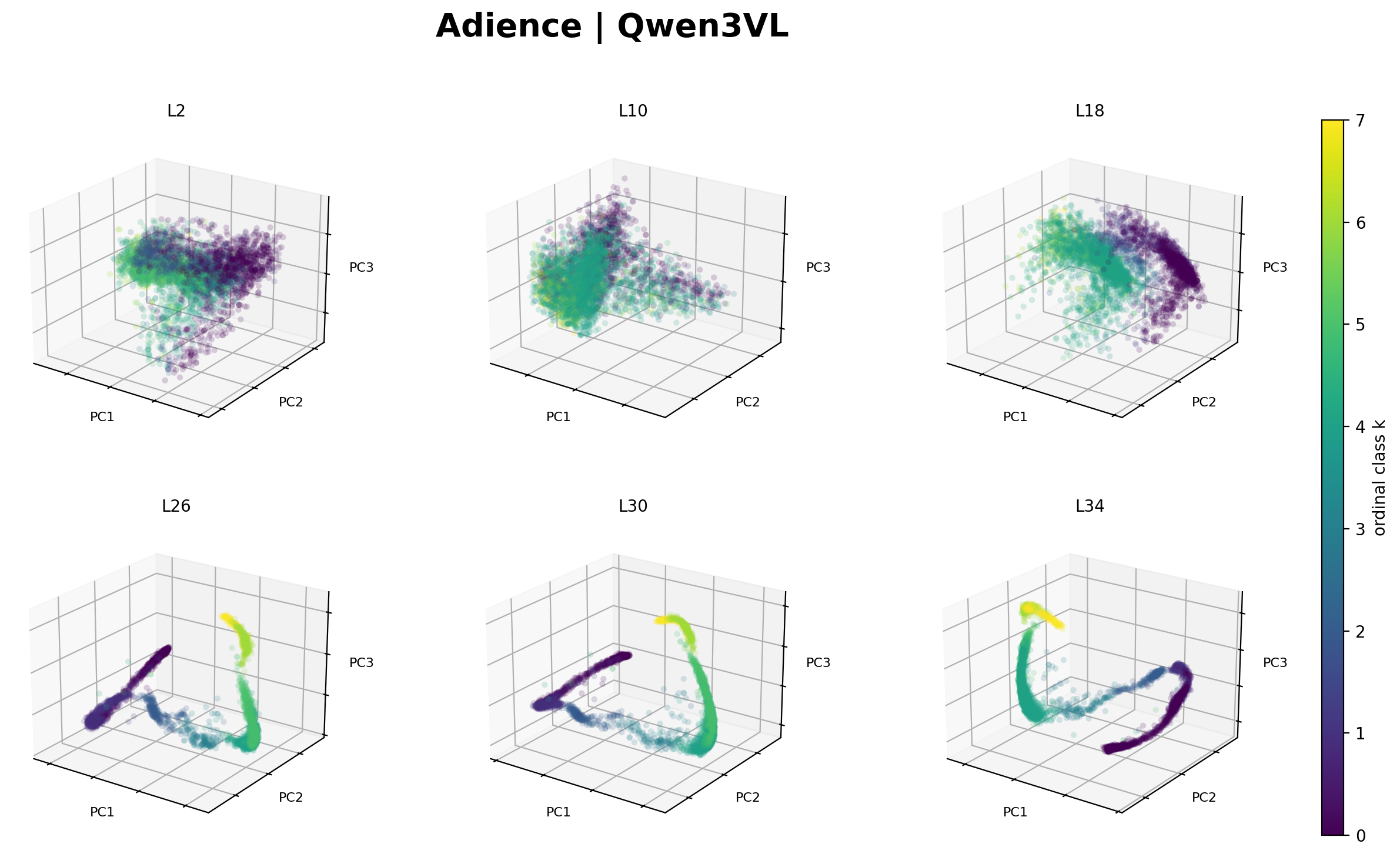}
\caption{Layer-wise 3D PCA of last-token hidden states on Adience under Qwen3-VL. Points are coloured by the eight age groups; the ordinal arc emerges in the middle-to-late decoder layers.}
\label{fig:pca-adience-qwen3vl}
\end{figure}

\begin{figure}[!htbp]
\centering
\includegraphics[width=0.85\linewidth]{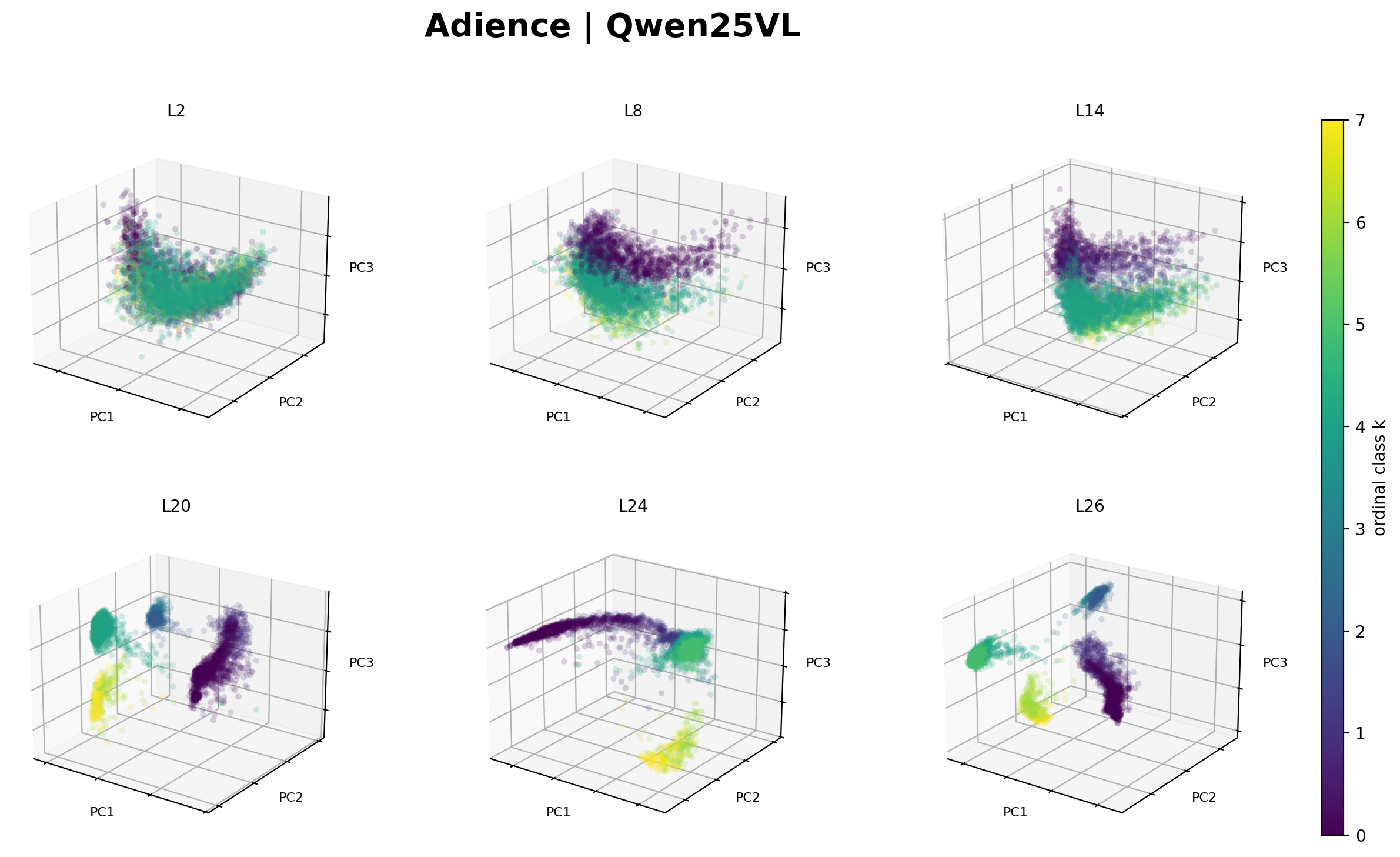}
\caption{Layer-wise 3D PCA on Adience under Qwen2.5-VL.}
\label{fig:pca-adience-qwen25vl}
\end{figure}

\begin{figure}[!htbp]
\centering
\includegraphics[width=0.85\linewidth]{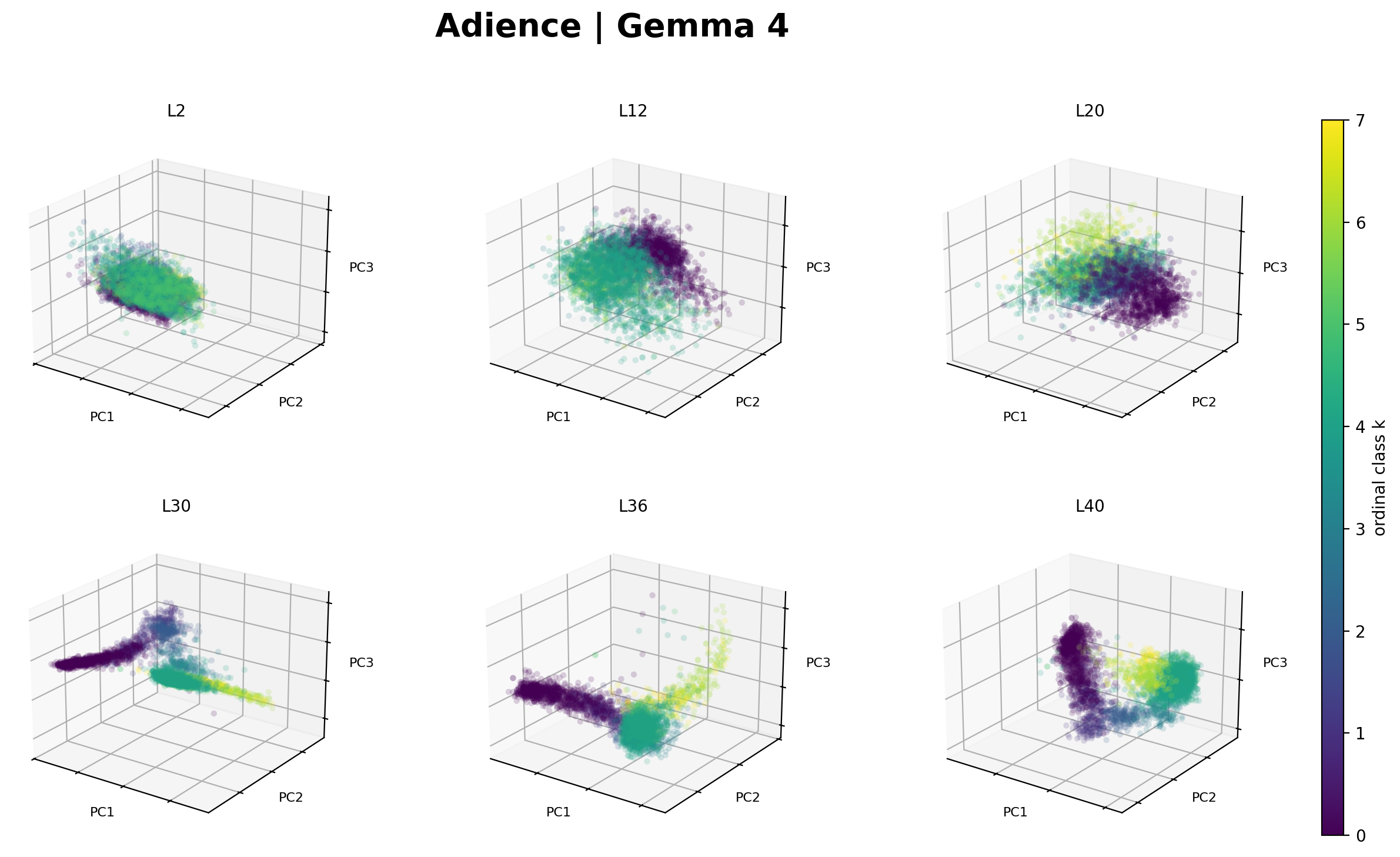}
\caption{Layer-wise 3D PCA on Adience under Gemma-4.}
\label{fig:pca-adience-gemma4}
\end{figure}

\begin{figure}[!htbp]
\centering
\includegraphics[width=0.85\linewidth]{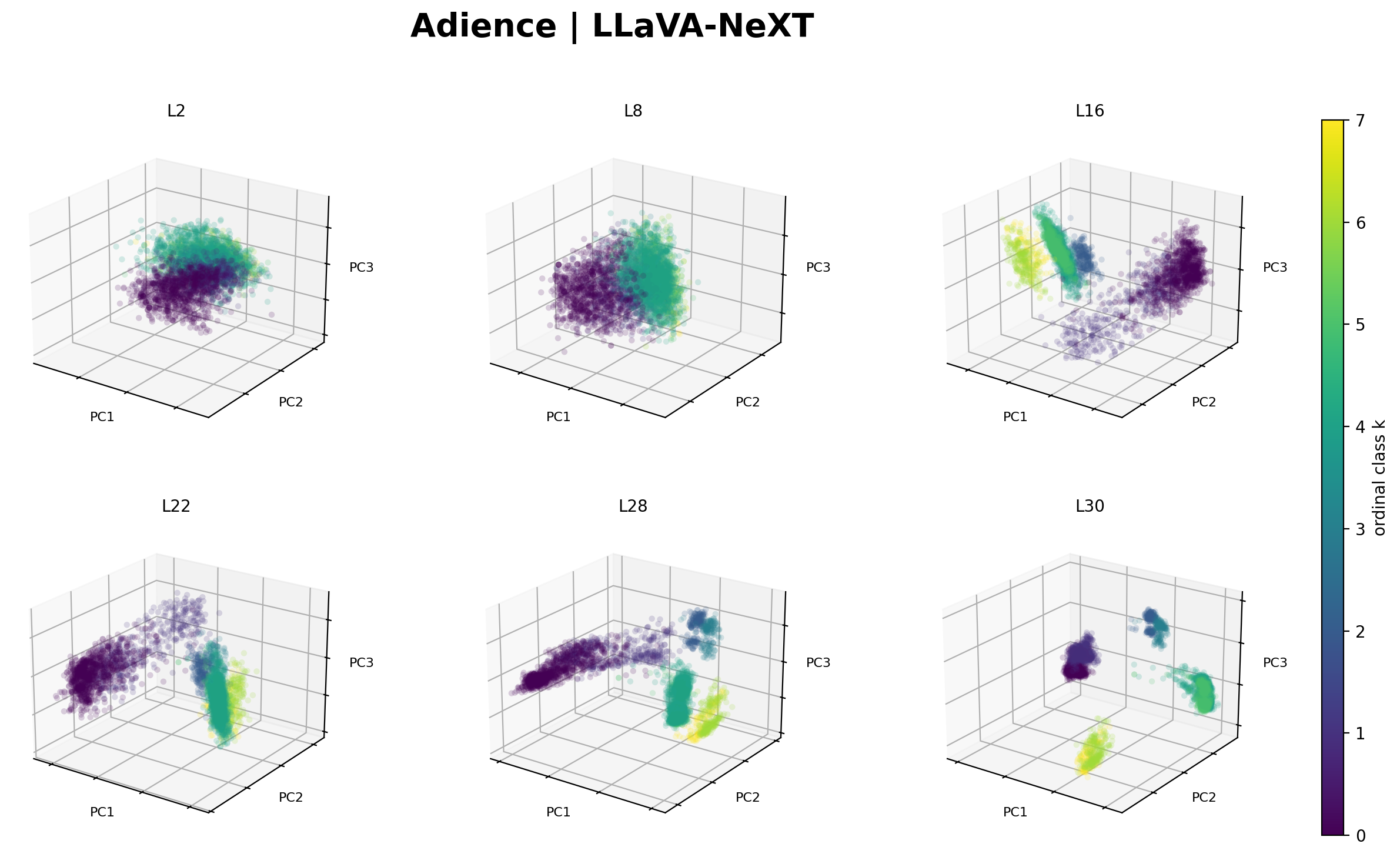}
\caption{Layer-wise 3D PCA on Adience under LLaVA-NeXT.}
\label{fig:pca-adience-llava-next}
\end{figure}

\begin{figure}[!htbp]
\centering
\includegraphics[width=0.85\linewidth]{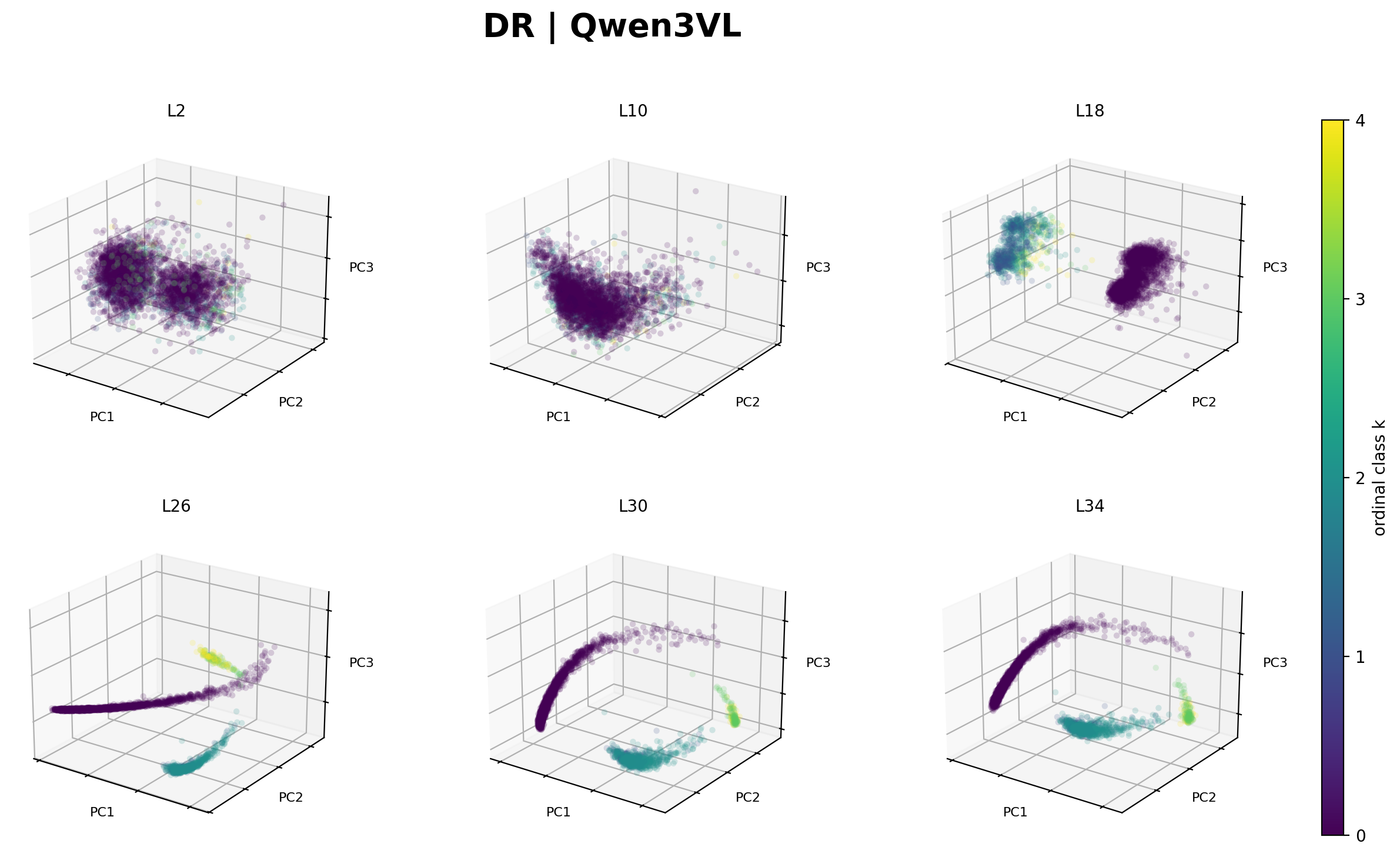}
\caption{Layer-wise 3D PCA on DR under Qwen3-VL. Points are coloured by the five ICDR severity grades.}
\label{fig:pca-dr-qwen3vl}
\end{figure}

\begin{figure}[!htbp]
\centering
\includegraphics[width=0.85\linewidth]{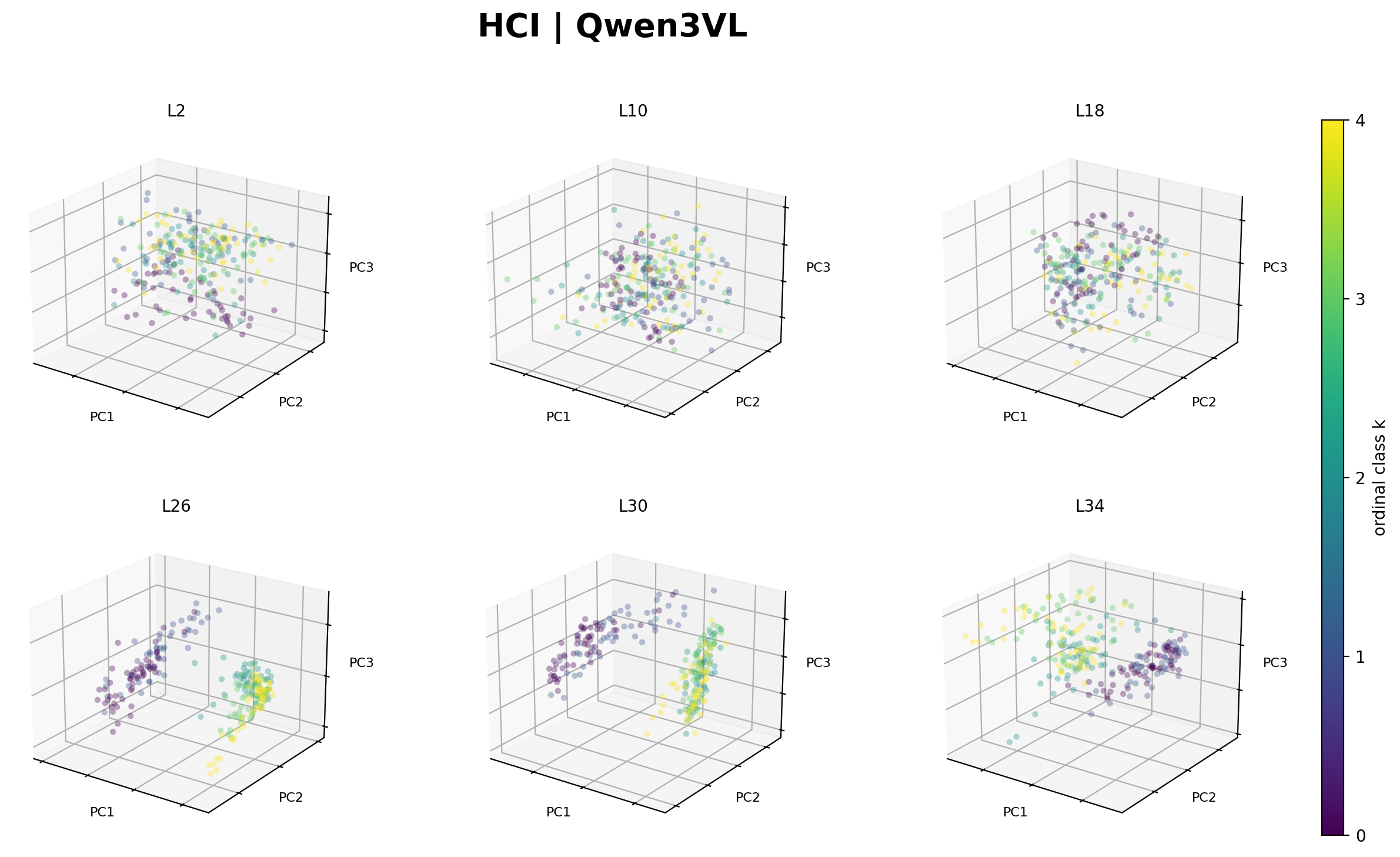}
\caption{Layer-wise 3D PCA on HCI under Qwen3-VL. Points are coloured by the five decade classes.}
\label{fig:pca-hci-qwen3vl}
\end{figure}

\begin{figure}[!htbp]
\centering
\includegraphics[width=0.85\linewidth]{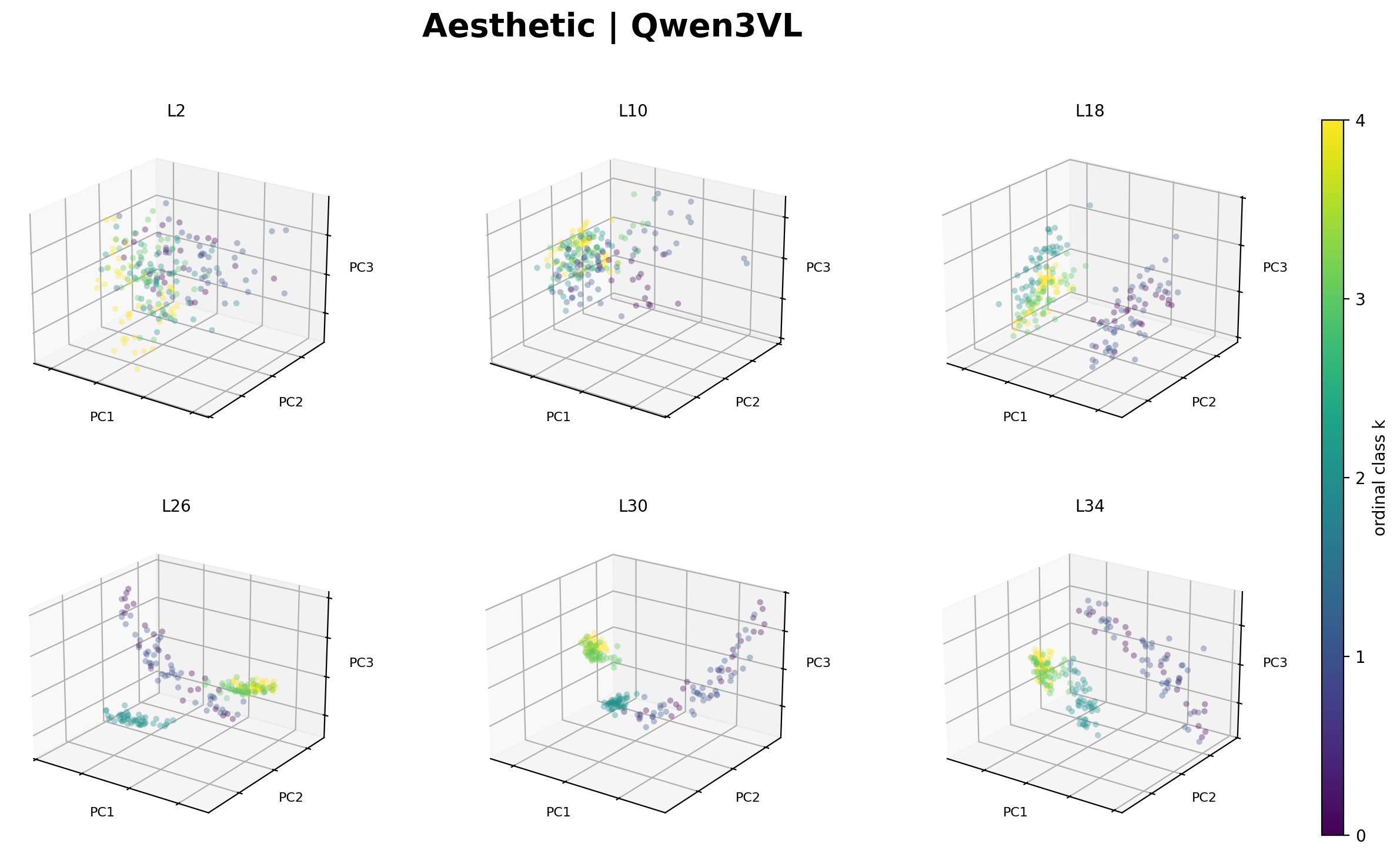}
\caption{Layer-wise 3D PCA on Aesthetic under Qwen3-VL. Points are coloured by the five aesthetic grades.}
\label{fig:pca-aesthetic-qwen3vl}
\end{figure}

We project the last-token hidden states onto their top three principal components at representative decoder layers and colour the points by ordinal class. Figures~\ref{fig:pca-adience-qwen3vl}--\ref{fig:pca-aesthetic-qwen3vl} show the resulting layer-wise structure: an ordinal arc emerges in the middle-to-late decoder layers, consistent with the per-layer SRCC peaks reported above.

\FloatBarrier

\subsection{Main results with five-fold standard deviations}
\label{app:main-results-detailed}

Table~\ref{tab:main-results-detailed} reports the main benchmark with mean~$\pm$~std over five folds for Adience, DR, and HCI; Aesthetic uses one class-balanced fold.

\begin{table*}[!htbp]
\centering\small
\resizebox{\textwidth}{!}{%
\begin{tabular}{llcccccccc}
\toprule
& & \multicolumn{2}{c}{Adience} & \multicolumn{2}{c}{DR} & \multicolumn{2}{c}{HCI} & \multicolumn{2}{c}{Aesthetic} \\
\cmidrule(lr){3-4}\cmidrule(lr){5-6}\cmidrule(lr){7-8}\cmidrule(lr){9-10}
Backbone & Method & ACC$\uparrow$ & MAE$\downarrow$ & ACC$\uparrow$ & MAE$\downarrow$ & ACC$\uparrow$ & MAE$\downarrow$ & ACC$\uparrow$ & MAE$\downarrow$ \\
\midrule
\multirow{4}{*}{Qwen2.5-VL}
 & Naive Prompt        & 0.5722 {\tiny$\pm$0.0168} & 0.5442 {\tiny$\pm$0.0644} & 0.1564 {\tiny$\pm$0.0058} & 1.5670 {\tiny$\pm$0.0118} & 0.2788 {\tiny$\pm$0.0027} & 0.9520 {\tiny$\pm$0.0044} & 0.3723 & 0.7721 \\
 & CAA                 & 0.6284 {\tiny$\pm$0.1641} & 0.3956 {\tiny$\pm$0.3129} & 0.7022 {\tiny$\pm$0.0142} & 0.3574 {\tiny$\pm$0.0211} & 0.3381 {\tiny$\pm$0.0396} & 0.8167 {\tiny$\pm$0.0882} & 0.2453 & 0.9283 \\
 & OrderChain          & 0.7650 {\tiny$\pm$0.0344} & 0.2823 {\tiny$\pm$0.0127} & 0.8700 {\tiny$\pm$0.0098} & 0.1400 {\tiny$\pm$0.0016} & 0.6528 {\tiny$\pm$0.0423} & 0.3473 {\tiny$\pm$0.0223} & 0.7053 & 0.2947 \\
 & \textbf{OLA (ours)} & \textbf{0.8088} {\tiny$\pm$0.0156} & \textbf{0.2327} {\tiny$\pm$0.0045} & \textbf{0.9201} {\tiny$\pm$0.0040} & \textbf{0.0799} {\tiny$\pm$0.0004} & \textbf{0.6848} {\tiny$\pm$0.0281} & \textbf{0.3153} {\tiny$\pm$0.0132} & \textbf{0.6962} & \textbf{0.3039} \\
\midrule
\multirow{4}{*}{Qwen3-VL}
 & Naive Prompt        & 0.5805 {\tiny$\pm$0.0340} & 0.4011 {\tiny$\pm$0.0299} & 0.6882 {\tiny$\pm$0.0121} & 0.5704 {\tiny$\pm$0.0229} & 0.3281 {\tiny$\pm$0.0201} & 0.9526 {\tiny$\pm$0.0202} & 0.3969 & 0.8588 \\
 & CAA                 & 0.6484 {\tiny$\pm$0.1641} & 0.3743 {\tiny$\pm$0.2961} & 0.7222 {\tiny$\pm$0.0142} & 0.3334 {\tiny$\pm$0.0197} & 0.3582 {\tiny$\pm$0.0396} & 0.7921 {\tiny$\pm$0.0852} & 0.2653 & 0.9037 \\
 & OrderChain          & 0.7898 {\tiny$\pm$0.0252} & 0.2103 {\tiny$\pm$0.0067} & 0.8950 {\tiny$\pm$0.0043} & 0.1100 {\tiny$\pm$0.0005} & 0.6552 {\tiny$\pm$0.0315} & 0.5301 {\tiny$\pm$0.0251} & 0.7101 & 0.2942 \\
 & \textbf{OLA (ours)} & \textbf{0.8059} {\tiny$\pm$0.0206} & \textbf{0.2068} {\tiny$\pm$0.0053} & \textbf{0.9255} {\tiny$\pm$0.0022} & \textbf{0.0780} {\tiny$\pm$0.0002} & \textbf{0.7573} {\tiny$\pm$0.0409} & \textbf{0.2602} {\tiny$\pm$0.0141} & \textbf{0.6977} & \textbf{0.3024} \\
\midrule
\multirow{4}{*}{Gemma-4}
 & Naive Prompt        & 0.5714 {\tiny$\pm$0.0179} & 0.4890 {\tiny$\pm$0.0314} & 0.1370 {\tiny$\pm$0.0046} & 1.4291 {\tiny$\pm$0.0227} & 0.2444 {\tiny$\pm$0.0161} & 0.9958 {\tiny$\pm$0.0256} & 0.2138 & 0.9627 \\
 & CAA                 & 0.5743 {\tiny$\pm$0.0461} & 0.4373 {\tiny$\pm$0.1036} & 0.7441 {\tiny$\pm$0.0165} & 0.3071 {\tiny$\pm$0.0126} & 0.3191 {\tiny$\pm$0.0054} & 0.8957 {\tiny$\pm$0.0519} & 0.3412 & 0.8156 \\
 & OrderChain          & 0.7548 {\tiny$\pm$0.0312} & 0.2453 {\tiny$\pm$0.0093} & 0.8800 {\tiny$\pm$0.0086} & 0.1450 {\tiny$\pm$0.0014} & 0.5804 {\tiny$\pm$0.0258} & 0.5100 {\tiny$\pm$0.0227} & 0.6805 & 0.3196 \\
 & \textbf{OLA (ours)} & \textbf{0.7602} {\tiny$\pm$0.0649} & \textbf{0.2399} {\tiny$\pm$0.0027} & \textbf{0.9219} {\tiny$\pm$0.0059} & \textbf{0.0782} {\tiny$\pm$0.0005} & \textbf{0.7329} {\tiny$\pm$0.0250} & \textbf{0.2700} {\tiny$\pm$0.0092} & \textbf{0.7005} & \textbf{0.2995} \\
\midrule
\multirow{4}{*}{LLaVA-NeXT}
 & Naive Prompt        & 0.4501 {\tiny$\pm$0.1295} & 0.7659 {\tiny$\pm$0.1685} & 0.1478 {\tiny$\pm$0.0101} & 1.8911 {\tiny$\pm$0.0277} & 0.2156 {\tiny$\pm$0.0221} & 1.0430 {\tiny$\pm$0.0251} & 0.3069 & 0.9842 \\
 & CAA                 & 0.6350 {\tiny$\pm$0.0221} & 0.3651 {\tiny$\pm$0.0283} & 0.2050 {\tiny$\pm$0.0245} & 1.1861 {\tiny$\pm$0.1056} & 0.3422 {\tiny$\pm$0.0266} & 0.9663 {\tiny$\pm$0.1768} & 0.3654 & 0.7805 \\
 & OrderChain          & 0.6821 {\tiny$\pm$0.0506} & 0.3671 {\tiny$\pm$0.0272} & 0.8392 {\tiny$\pm$0.0161} & 0.2214 {\tiny$\pm$0.0042} & 0.6221 {\tiny$\pm$0.0371} & 0.4384 {\tiny$\pm$0.0262} & 0.6957 & 0.3243 \\
 & \textbf{OLA (ours)} & \textbf{0.7910} {\tiny$\pm$0.0410} & \textbf{0.2293} {\tiny$\pm$0.0119} & \textbf{0.9242} {\tiny$\pm$0.0083} & \textbf{0.0759} {\tiny$\pm$0.0006} & \textbf{0.7232} {\tiny$\pm$0.0553} & \textbf{0.3083} {\tiny$\pm$0.0236} & \textbf{0.7189} & \textbf{0.2812} \\
\bottomrule
\end{tabular}}
\caption{Main results on four ordinal benchmarks under four MLLM backbones, with mean~$\pm$~std over five folds (Adience, DR, HCI); Aesthetic uses one class-balanced fold.}
\label{tab:main-results-detailed}
\end{table*}

\subsection{Ablation studies with five-fold standard deviations}
\label{app:ablation-results-detailed}

Table~\ref{tab:ablation-results-detailed} reports the full ablation studies under four MLLM backbones on all four datasets. Eight methods are compared per backbone-dataset cell: Capacity Probe, Probe Only, Probe Steering, Single-Layer Logit, Designed Prompt, OL~+~Naive, OLA-Offline, and OLA-Online.

\begin{table*}[!htbp]
\centering\small
\resizebox{\textwidth}{!}{%
\begin{tabular}{llcccccccc}
\toprule
& & \multicolumn{2}{c}{Adience} & \multicolumn{2}{c}{DR} & \multicolumn{2}{c}{HCI} & \multicolumn{2}{c}{Aesthetic} \\
\cmidrule(lr){3-4}\cmidrule(lr){5-6}\cmidrule(lr){7-8}\cmidrule(lr){9-10}
Backbone & Method & ACC$\uparrow$ & MAE$\downarrow$ & ACC$\uparrow$ & MAE$\downarrow$ & ACC$\uparrow$ & MAE$\downarrow$ & ACC$\uparrow$ & MAE$\downarrow$ \\
\midrule
\multirow{8}{*}{Qwen2.5-VL}
 & Capacity Probe       & 0.7923 {\tiny$\pm$0.0250} & 0.2721 {\tiny$\pm$0.0086} & 0.8941 {\tiny$\pm$0.0178} & 0.0821 {\tiny$\pm$0.0016} & 0.6161 {\tiny$\pm$0.0356} & 0.4009 {\tiny$\pm$0.0232} & 0.6871 & 0.3291 \\
 & Probe Only           & 0.7800 {\tiny$\pm$0.0286} & 0.2667 {\tiny$\pm$0.0098} & 0.9100 {\tiny$\pm$0.0106} & 0.0950 {\tiny$\pm$0.0011} & 0.6230 {\tiny$\pm$0.0251} & 0.3904 {\tiny$\pm$0.0157} & 0.6700 & 0.3500 \\
 & Probe Steering       & 0.6496 {\tiny$\pm$0.1851} & 0.3639 {\tiny$\pm$0.2881} & 0.7246 {\tiny$\pm$0.0071} & 0.3306 {\tiny$\pm$0.0196} & 0.3864 {\tiny$\pm$0.0619} & 0.7488 {\tiny$\pm$0.0806} & 0.4293 & 0.6658 \\
 & Single-Layer Logit   & 0.7150 {\tiny$\pm$0.0215} & 0.3050 {\tiny$\pm$0.0092} & 0.8950 {\tiny$\pm$0.0086} & 0.1100 {\tiny$\pm$0.0011} & 0.6200 {\tiny$\pm$0.0258} & 0.4100 {\tiny$\pm$0.0171} & 0.5300 & 0.4701 \\
 & Designed Prompt      & 0.6012 {\tiny$\pm$0.0280} & 0.3989 {\tiny$\pm$0.0180} & 0.8300 {\tiny$\pm$0.0054} & 0.2270 {\tiny$\pm$0.0015} & 0.4825 {\tiny$\pm$0.0248} & 0.6104 {\tiny$\pm$0.0314} & 0.4541 & 0.6812 \\
 & OL + Naive           & 0.6521 {\tiny$\pm$0.0626} & 0.3653 {\tiny$\pm$0.0250} & 0.5341 {\tiny$\pm$0.1525} & 0.6532 {\tiny$\pm$0.0760} & 0.5125 {\tiny$\pm$0.0531} & 0.5645 {\tiny$\pm$0.0314} & 0.5153 & 0.5080 \\
 & OLA-Offline          & 0.7918 {\tiny$\pm$0.0107} & 0.2593 {\tiny$\pm$0.0036} & 0.8722 {\tiny$\pm$0.0274} & 0.0963 {\tiny$\pm$0.0031} & 0.6252 {\tiny$\pm$0.0192} & 0.3821 {\tiny$\pm$0.0117} & 0.6867 & 0.3227 \\
 & \textbf{OLA-Online}  & \textbf{0.8088} {\tiny$\pm$0.0156} & \textbf{0.2327} {\tiny$\pm$0.0045} & \textbf{0.9201} {\tiny$\pm$0.0040} & \textbf{0.0799} {\tiny$\pm$0.0004} & \textbf{0.6848} {\tiny$\pm$0.0281} & \textbf{0.3153} {\tiny$\pm$0.0132} & \textbf{0.6962} & \textbf{0.3039} \\
\midrule
\multirow{8}{*}{Qwen3-VL}
 & Capacity Probe       & 0.7809 {\tiny$\pm$0.0316} & 0.2198 {\tiny$\pm$0.0089} & 0.9126 {\tiny$\pm$0.0102} & 0.0824 {\tiny$\pm$0.0009} & 0.7160 {\tiny$\pm$0.0562} & 0.2932 {\tiny$\pm$0.0231} & 0.6461 & 0.3740 \\
 & Probe Only           & 0.7851 {\tiny$\pm$0.0257} & 0.2300 {\tiny$\pm$0.0075} & 0.9107 {\tiny$\pm$0.0097} & 0.0894 {\tiny$\pm$0.0009} & 0.6900 {\tiny$\pm$0.0258} & 0.3564 {\tiny$\pm$0.0133} & 0.6200 & 0.4400 \\
 & Probe Steering       & 0.6696 {\tiny$\pm$0.1851} & 0.3431 {\tiny$\pm$0.2716} & 0.7446 {\tiny$\pm$0.0070} & 0.3065 {\tiny$\pm$0.0181} & 0.4064 {\tiny$\pm$0.0619} & 0.7244 {\tiny$\pm$0.0779} & 0.4493 & 0.6425 \\
 & Single-Layer Logit   & 0.7750 {\tiny$\pm$0.0172} & 0.2800 {\tiny$\pm$0.0062} & 0.9080 {\tiny$\pm$0.0034} & 0.0950 {\tiny$\pm$0.0004} & 0.6300 {\tiny$\pm$0.0258} & 0.4262 {\tiny$\pm$0.0252} & 0.5900 & 0.4700 \\
 & Designed Prompt      & 0.6274 {\tiny$\pm$0.0212} & 0.4009 {\tiny$\pm$0.0135} & 0.8398 {\tiny$\pm$0.0114} & 0.1755 {\tiny$\pm$0.0024} & 0.3742 {\tiny$\pm$0.0370} & 0.8511 {\tiny$\pm$0.0252} & 0.4300 & 0.8116 \\
 & OL + Naive           & 0.6381 {\tiny$\pm$0.0539} & 0.3872 {\tiny$\pm$0.0220} & 0.7941 {\tiny$\pm$0.0724} & 0.2366 {\tiny$\pm$0.0362} & 0.4215 {\tiny$\pm$0.0852} & 0.7115 {\tiny$\pm$0.0252} & 0.5248 & 0.5493 \\
 & OLA-Offline          & 0.7998 {\tiny$\pm$0.0378} & 0.2102 {\tiny$\pm$0.0099} & 0.9163 {\tiny$\pm$0.0120} & 0.0797 {\tiny$\pm$0.0011} & 0.7260 {\tiny$\pm$0.0413} & 0.2816 {\tiny$\pm$0.0161} & 0.6528 & 0.3692 \\
 & \textbf{OLA-Online}  & \textbf{0.8059} {\tiny$\pm$0.0206} & \textbf{0.2068} {\tiny$\pm$0.0053} & \textbf{0.9255} {\tiny$\pm$0.0022} & \textbf{0.0780} {\tiny$\pm$0.0002} & \textbf{0.7573} {\tiny$\pm$0.0409} & \textbf{0.2602} {\tiny$\pm$0.0141} & \textbf{0.6977} & \textbf{0.3024} \\
\midrule
\multirow{8}{*}{Gemma-4}
 & Capacity Probe       & 0.7324 {\tiny$\pm$0.0201} & 0.2605 {\tiny$\pm$0.0071} & 0.9130 {\tiny$\pm$0.0119} & 0.0822 {\tiny$\pm$0.0011} & 0.7294 {\tiny$\pm$0.0196} & 0.3152 {\tiny$\pm$0.0086} & 0.6817 & 0.3159 \\
 & Probe Only           & 0.7350 {\tiny$\pm$0.0215} & 0.2750 {\tiny$\pm$0.0080} & 0.9050 {\tiny$\pm$0.0060} & 0.1080 {\tiny$\pm$0.0007} & 0.7013 {\tiny$\pm$0.0200} & 0.3600 {\tiny$\pm$0.0103} & 0.6800 & 0.3300 \\
 & Probe Steering       & 0.6229 {\tiny$\pm$0.0677} & 0.3872 {\tiny$\pm$0.0916} & 0.7854 {\tiny$\pm$0.0087} & 0.2575 {\tiny$\pm$0.0106} & 0.3220 {\tiny$\pm$0.0028} & 0.8800 {\tiny$\pm$0.0509} & 0.3333 & 0.8261 \\
 & Single-Layer Logit   & 0.7250 {\tiny$\pm$0.0215} & 0.2900 {\tiny$\pm$0.0086} & 0.8950 {\tiny$\pm$0.0043} & 0.1200 {\tiny$\pm$0.0006} & 0.6500 {\tiny$\pm$0.0271} & 0.4100 {\tiny$\pm$0.0171} & 0.6200 & 0.4300 \\
 & Designed Prompt      & 0.6915 {\tiny$\pm$0.0460} & 0.3086 {\tiny$\pm$0.0195} & 0.8282 {\tiny$\pm$0.0150} & 0.1790 {\tiny$\pm$0.0032} & 0.2920 {\tiny$\pm$0.0275} & 0.9024 {\tiny$\pm$0.0850} & 0.3478 & 0.8551 \\
 & OL + Naive           & 0.6835 {\tiny$\pm$0.0638} & 0.3197 {\tiny$\pm$0.0240} & 0.4925 {\tiny$\pm$0.1242} & 0.7049 {\tiny$\pm$0.0621} & 0.3924 {\tiny$\pm$0.0742} & 0.7722 {\tiny$\pm$0.0500} & 0.5625 & 0.5119 \\
 & OLA-Offline          & 0.7497 {\tiny$\pm$0.0147} & 0.2305 {\tiny$\pm$0.0046} & 0.9197 {\tiny$\pm$0.0064} & 0.0790 {\tiny$\pm$0.0006} & 0.7144 {\tiny$\pm$0.0274} & 0.3448 {\tiny$\pm$0.0132} & 0.6957 & 0.3043 \\
 & \textbf{OLA-Online}  & \textbf{0.7602} {\tiny$\pm$0.0649} & \textbf{0.2399} {\tiny$\pm$0.0027} & \textbf{0.9219} {\tiny$\pm$0.0059} & \textbf{0.0782} {\tiny$\pm$0.0005} & \textbf{0.7329} {\tiny$\pm$0.0250} & \textbf{0.2700} {\tiny$\pm$0.0092} & \textbf{0.7005} & \textbf{0.2995} \\
\midrule
\multirow{8}{*}{LLaVA-NeXT}
 & Capacity Probe       & 0.7762 {\tiny$\pm$0.0586} & 0.2847 {\tiny$\pm$0.0216} & 0.8929 {\tiny$\pm$0.0345} & 0.0771 {\tiny$\pm$0.0031} & 0.7028 {\tiny$\pm$0.0132} & 0.3312 {\tiny$\pm$0.0062} & 0.7031 & 0.2995 \\
 & Probe Only           & 0.7550 {\tiny$\pm$0.0350} & 0.3021 {\tiny$\pm$0.0140} & 0.8400 {\tiny$\pm$0.0100} & 0.1750 {\tiny$\pm$0.0021} & 0.7023 {\tiny$\pm$0.0268} & 0.3500 {\tiny$\pm$0.0134} & 0.7053 & 0.3012 \\
 & Probe Steering       & 0.6217 {\tiny$\pm$0.0281} & 0.3808 {\tiny$\pm$0.0297} & 0.1893 {\tiny$\pm$0.0024} & 1.2113 {\tiny$\pm$0.1077} & 0.3355 {\tiny$\pm$0.0566} & 0.9788 {\tiny$\pm$0.1789} & 0.5266 & 0.5990 \\
 & Single-Layer Logit   & 0.6902 {\tiny$\pm$0.0248} & 0.3150 {\tiny$\pm$0.0113} & 0.8213 {\tiny$\pm$0.0208} & 0.2574 {\tiny$\pm$0.0065} & 0.6042 {\tiny$\pm$0.0302} & 0.4561 {\tiny$\pm$0.0253} & 0.5300 & 0.5500 \\
 & Designed Prompt      & 0.6244 {\tiny$\pm$0.0180} & 0.3813 {\tiny$\pm$0.0110} & 0.7982 {\tiny$\pm$0.0174} & 0.2590 {\tiny$\pm$0.0056} & 0.3523 {\tiny$\pm$0.0121} & 0.8861 {\tiny$\pm$0.0304} & 0.3671 & 0.8986 \\
 & OL + Naive           & 0.5934 {\tiny$\pm$0.0214} & 0.4269 {\tiny$\pm$0.0150} & 0.6835 {\tiny$\pm$0.1142} & 0.4598 {\tiny$\pm$0.0571} & 0.3676 {\tiny$\pm$0.0519} & 0.8569 {\tiny$\pm$0.0304} & 0.4924 & 0.6589 \\
 & OLA-Offline          & 0.7890 {\tiny$\pm$0.0524} & 0.2610 {\tiny$\pm$0.0173} & 0.9088 {\tiny$\pm$0.0138} & 0.0712 {\tiny$\pm$0.0011} & 0.7180 {\tiny$\pm$0.0620} & 0.3288 {\tiny$\pm$0.0284} & 0.7146 & 0.2754 \\
 & \textbf{OLA-Online}  & \textbf{0.7910} {\tiny$\pm$0.0410} & \textbf{0.2293} {\tiny$\pm$0.0119} & \textbf{0.9242} {\tiny$\pm$0.0083} & \textbf{0.0759} {\tiny$\pm$0.0006} & \textbf{0.7232} {\tiny$\pm$0.0553} & \textbf{0.3083} {\tiny$\pm$0.0236} & \textbf{0.7189} & \textbf{0.2812} \\
\bottomrule
\end{tabular}}
\caption{Ablation studies on four ordinal benchmarks under four MLLM backbones (mean~$\pm$~std over five folds for Adience, DR, HCI; Aesthetic uses one class-balanced fold).}
\label{tab:ablation-results-detailed}
\end{table*}

\clearpage

\section{Pseudocode}
\label{app:pseudocode}

Algorithms~\ref{alg:lens-learning} and~\ref{alg:online-alignment} give the pseudocode for the two OLA phases that match the method overview in Figure~\ref{fig:method}; lines that are pure I/O or device management are omitted for brevity.

\begin{algorithm}[H]
\caption{Multi-layer Ordinal Lens Learning}
\label{alg:lens-learning}
\small
\begin{algorithmic}[1]
\Require frozen MLLM $M$, dataset $\mathcal{D}$, designed prompt $P$, layer set $\mathcal{L}$, digit-row matrix $W_S$ (frozen)
\Ensure lens parameters $\{A_\ell, b_\ell\}_{\ell\in\mathcal{L}}$, fusion logits $\alpha$, cached $P^{\mathrm{OLA}}(\cdot\mid x)$
\Statex
\Statex \textit{// one forward pass per sample, cache last-token hidden}
\State register forward hooks on every layer $\ell\in\mathcal{L}$
\For{each $x\in\mathcal{D}$}
  \State $\text{out}\gets M.\textsc{forward}(P, \text{image}(x))$
  \State $H[x,\ell]\gets$ captured last-token hidden at layer $\ell$, for $\ell\in\mathcal{L}$
\EndFor
\Statex
\Statex \textit{// Stage A: per-layer ordinal lens (independent training)}
\For{each $\ell\in\mathcal{L}$}
  \State $(\mu_\ell,\sigma_\ell)\gets$ mean and clamped std of $H[:,\ell]$ on train split
  \State initialise $A_\ell\gets I+U_\ell V_\ell^\top$ with $U_\ell\sim\mathcal{N}(0,0.02)$, $V_\ell\gets 0$; $b_\ell\gets 0$
  \For{$\text{epoch}=1$ to $100$}
    \For{minibatch $B$}
      \State $\tilde h\gets (H[B,\ell]-\mu_\ell)/\sigma_\ell$
      \State $\hat z_\ell\gets W_S(A_\ell\tilde h+b_\ell)$
      \State $\mathrm{CE}(\mathrm{softmax}(\hat z_\ell), y[B]).\textsc{backward}()$; clip-norm $1.0$; AdamW step
    \EndFor
    \State track val accuracy; keep best-epoch $(A_\ell, b_\ell)$
  \EndFor
\EndFor
\Statex
\Statex \textit{// Stage B: softmax fusion over $K$ layers}
\State freeze all $\{A_\ell, b_\ell\}$; $\alpha\gets 0$; require\_grad$(\alpha)$
\For{$\text{epoch}=1$ to $30$}
  \For{minibatch $B$}
    \State $\hat z_\ell(B)\gets W_S(A_\ell\tilde h_\ell(B)+b_\ell)$ \ \ for $\ell\in\mathcal{L}$
    \State $a\gets\mathrm{softmax}(\alpha)$
    \State $z^{\mathrm{OLA}}\gets\sum_{\ell\in\mathcal{L}} a_\ell\,\hat z_\ell(B)$
    \State $\mathrm{CE}(\mathrm{softmax}(z^{\mathrm{OLA}}), y[B]).\textsc{backward}()$; AdamW step ($\alpha$ only)
  \EndFor
  \State track val accuracy; keep best-epoch $\alpha$
\EndFor
\Statex
\Statex \textit{// cache the instance-specific ordinal distribution}
\For{split in $\{\text{train}, \text{val}, \text{test}\}$}
  \State $P^{\mathrm{OLA}}(\cdot\mid x)\gets\mathrm{softmax}\bigl(\sum_\ell a_\ell\,\hat z_\ell(x)\bigr)$ for $x$ in split
\EndFor
\State \Return $\{A_\ell, b_\ell\}$, $\alpha$, $P^{\mathrm{OLA}}$
\end{algorithmic}
\end{algorithm}

\begin{algorithm}[H]
\caption{Inference-Time Ordinal Alignment}
\label{alg:online-alignment}
\small
\begin{algorithmic}[1]
\Require frozen MLLM $M$, designed prompt $P$, trained lens $\{A_\ell, b_\ell\}$ and $\alpha$, cached $P^{\mathrm{OLA}}(\cdot\mid x)$, base digit logits $L^{\mathrm{na}}[x]$ on val and test, grid $\Lambda$, digit-token ids $\mathcal{S}_{\mathrm{digit}}$, EOS ids $\mathcal{S}_{\mathrm{eos}}$
\Ensure $\lambda^\star$ and online-corrected predictions on test
\Statex
\Statex \textit{// 1. select $\lambda^\star$ on validation (offline grid sweep)}
\For{$\lambda\in\Lambda$}
  \For{$x$ in val}
    \State $Q(x)\gets\mathrm{softmax}(L^{\mathrm{na}}[x])$
    \State $\omega(x)\gets\max_c P^{\mathrm{OLA}}(c\mid x)$
    \State $\tilde L\gets L^{\mathrm{na}}[x]-\lambda\,(Q(x)-P^{\mathrm{OLA}}(\cdot\mid x))\,\omega(x)$
    \State $\hat y(x)\gets\arg\max_{c\in\mathcal{S}_{\mathrm{digit}}} \tilde L$
  \EndFor
  \State record $\mathrm{ACC}_{\mathrm{val}}(\lambda)$
\EndFor
\State $\lambda^\star\gets\arg\max_\lambda \mathrm{ACC}_{\mathrm{val}}(\lambda)$
\Statex
\Statex \textit{// 2. online generation on test (deterministic decoding)}
\State register forward hooks on every $\ell\in\mathcal{L}$ that write into a shared dict $H$
\For{$x$ in test}
  \State clear $H$ \Comment{per-sample reset}
  \State prefill $M$ with $(P, \text{image}(x))$ \Comment{hooks fill $H[\ell]$}
  \State $P^{\mathrm{OLA}}(\cdot\mid x)\gets\mathrm{softmax}\bigl(\sum_\ell a_\ell\, W_S(A_\ell H[\ell]_{\tilde{}}+b_\ell)\bigr)$
  \State $\omega(x)\gets\max_c P^{\mathrm{OLA}}(c\mid x)$
  \State greedily decode until reaching the answer slot \Comment{no correction}
  \State \textbf{at the answer slot:}
  \State \quad $L_t\gets M.\text{lm\_head\_logits}$
  \State \quad $Q_t(c)\gets\mathrm{softmax}(L_t[\mathcal{S}_{\mathrm{digit}}])$
  \State \quad $L_t[\mathcal{S}_{\mathrm{digit}}]\gets L_t[\mathcal{S}_{\mathrm{digit}}]-\lambda^\star\,(Q_t-P^{\mathrm{OLA}}(\cdot\mid x))\,\omega(x)$
  \State \quad emit $\arg\max(L_t)$ as the ordinal prediction
  \State continue greedy decoding until EOS \Comment{no correction}
\EndFor
\end{algorithmic}
\end{algorithm}

\section{Related Work}
\label{app:related_work}

\subsection{Discriminative Ordinal Regression}
Ordinal regression treats prediction as classification over ordered labels, where the distance between predicted and true classes is meaningful. Early work formulated this as a neural ranking problem \citep{25} and decomposed the $C$-class ordinal task into $C{-}1$ binary subproblems \citep{2}. Subsequent work introduced rank consistency through shared bias terms (CORAL, \citealp{1}) and conditional probabilities (CORN, \citealp{5}). Other approaches encode ordinal structure through soft labels that smooth class boundaries \citep{8}, label diversity that decorrelates probe directions \citep{7}, or weak supervision for medical grading \citep{3}. Domain-specific work includes face-age estimation \citep{9} and neural image assessment \citep{4}. Number-aware vision-language models such as NumCLIP \citep{26} and OrdinalCLIP \citep{6} introduce contrastive objectives that preserve numerical and ordinal relations in visual representations, and sequence-based formulations such as Ord2Seq \citep{27} reduce ordinal classification to autoregressive sequence prediction. Our work extends this diagnostic perspective to frozen MLLM hidden states and the digit-token interface, rather than training a discriminative ordinal head.

\subsection{Ordinal Prediction With MLLMs}
Recent work applies MLLMs to ordinal recognition by reformulating ordered labels as digit-token generation. OrderChain is the closest baseline: it combines task-specific prompting with LoRA fine-tuning on age, aesthetic, and medical grading benchmarks \citep{28}. A separate line of work audits MLLMs as ordinal raters in clinical settings, reporting central-tendency bias on clinical ordinal scales \citep{19}. Rather than updating the backbone, OLA asks whether the frozen model already contains ordinal evidence and whether the output interface exposes it, separating backbone adaptation from inference-time output alignment. The MLLM backbones we study span the Qwen-VL family \citep{14, 13}, the Gemma open-weights family \citep{15}, and LLaVA-NeXT \citep{16}.

\subsection{Linear Probing and Representation Geometry}
Linear classifier probes \citep{17} measure how much task-relevant information a frozen representation linearly carries; structural probes \citep{18} extend this to relational structure such as syntax. The linear representation hypothesis \citep{21} further argues that concepts in large language models are approximately encoded as linear directions in the residual stream. Our diagnostic uses the same linear-readout family and additionally measures the spectral and row-space exposure of these directions through the unembedding matrix.

\subsection{Representation Steering and Parameter-Efficient Adaptation}
Representation steering methods control model behavior by adding or projecting directions in the residual stream, with directions estimated from contrastive examples or learned probes \citep{22}. While these methods influence generation via hidden-state directions, they still rely on the native LM head, which our diagnostics show can be unreliable for ordinal prediction when a strong hidden direction is weakly exposed by the digit-token rows. OLA therefore applies alignment at the output side, modifying only candidate digit-token logits with a signal derived from $W_S$-anchored ordinal lenses. Parameter-efficient adaptation such as LoRA \citep{23} provides an alternative route by training low-rank backbone updates; OLA differs by leaving the backbone frozen and training only output-side lens parameters.

\subsection{Lens-Based Interpretation and Spectral Output Filtering}
Logit-lens style methods map intermediate transformer states into vocabulary logits to analyze how predictions evolve across depth \citep{24}. Spectral analyses of the unembedding matrix identify structured filters and dark signals that the output interface imposes on residual-stream content \citep{20}. OLA uses the same readout family but trains task-specific ordinal lenses anchored on $W_S$ and uses their fused distribution for inference-time correction rather than interpretation only.

\end{document}